\documentclass[11pt]{article}

\usepackage[preprint]{acl}

\usepackage{times}
\usepackage{latexsym}

\usepackage[T1]{fontenc}
\usepackage[utf8]{inputenc}

\usepackage{microtype}

\usepackage{inconsolata}

\usepackage{graphicx}

\usepackage{hyperref}
\usepackage{url}

\usepackage{multirow}
\usepackage{color}
\usepackage{graphicx}
\usepackage{enumitem}
\usepackage{threeparttable}
\usepackage[export]{adjustbox}
\usepackage{caption}
\usepackage{subcaption}
\usepackage{booktabs}
\usepackage{colortbl}
\usepackage{algorithm}
\usepackage{algpseudocode}
\usepackage{geometry}
\usepackage{xcolor}
\usepackage{fontawesome}
\usepackage{pifont}
\usepackage{array}
\usepackage[nointegrals]{wasysym}
\usepackage{makecell}
\usepackage{amsmath}
\usepackage{amsfonts}

\newcommand{\system}{\textsc{SMAD}}

\title{\system: A Semantics-Guided Multi-Agent Framework for Unsupervised Time-Series Anomaly Detection}

\author{
 \textbf{Qinghua Liu\textsuperscript{1,2}}\thanks{Work was done during the internship at Visa Research.},
 \textbf{Junpeng Wang\textsuperscript{2}},
 \textbf{Mingzhi Hu\textsuperscript{2}},
 \textbf{Menghai Pan\textsuperscript{2}}, \\
 \textbf{Shubham Jain\textsuperscript{2}},
 \textbf{Yuzhong Chen\textsuperscript{2}},
 \textbf{Yan Zheng\textsuperscript{2}}, \\
 \textbf{Yiwei Cai\textsuperscript{2}},
 \textbf{John Paparrizos\textsuperscript{1,3}}
\\
 \textsuperscript{1}The Ohio State University,
 \textsuperscript{2}Visa Research,
 \textsuperscript{3}Aristotle University of Thessaloniki
}

\begin{document}
\maketitle

\begin{abstract}
Time-series anomaly detection (TSAD) lacks a one-size-fits-all detector for heterogeneous data and anomaly types. Existing automation is limited by implicit objectives, potentially unreliable historical knowledge, or high computational costs. More fundamentally, it selects \textit{how} to detect anomalies without explicitly defining \textit{what} to detect.
We propose \system, an LLM-based multi-agent framework that translates anomaly intent into executable, testable detection pipelines. Rather than selecting detectors against an implicit target, \system\ establishes a natural-language semantic contract grounded in measurable time-series properties, specifying target event patterns, expected score responses, and validation questions. This contract governs both detector construction and evaluation: LLM agents propose pipelines, while contract-matched synthetic interventions provide label-free evidence from the target series to assess semantic coverage and guide iterative self-refinement. A semantics-aware ensemble integrates validated candidates and attributes each alert to contributing detectors and supporting evidence.
On univariate and multivariate TSAD benchmarks, \system\ improves detection performance over the strongest baseline by $11.5\%$ and $9.6\%$, respectively, while outperforming general-purpose coding agents at $2$--$5\times$ lower language-model token cost. 
\end{abstract}

\section{Introduction}

Time-series data underpin cyber-physical infrastructure~\citep{osti_1478744,osti_1579323}, cloud services~\citep{calheiros2017effectiveness,bhanage2021infrastructure}, industrial equipment~\citep{susto2017anomaly,liu2018anomaly}, healthcare monitoring~\citep{li2020survey,chuah2007ecg}, and financial systems~\citep{huang2018codetect,hilal2022financial}. In these settings, time-series anomaly detection (TSAD) identifies rare but consequential deviations that may indicate failures, unsafe conditions, attacks, or unexpected operational changes. Reliable TSAD remains difficult~\citep{liu2024elephant,qiu2025tab} because real-world time series differ in temporal dependence, seasonality, trend, noise, and operating regime. Their anomalies are equally heterogeneous, ranging from isolated points and persistent level shifts to collective and context-dependent deviations~\citep{boniol2024dive}.

\begin{figure}[t]
    \centering
    \includegraphics[width=\columnwidth]{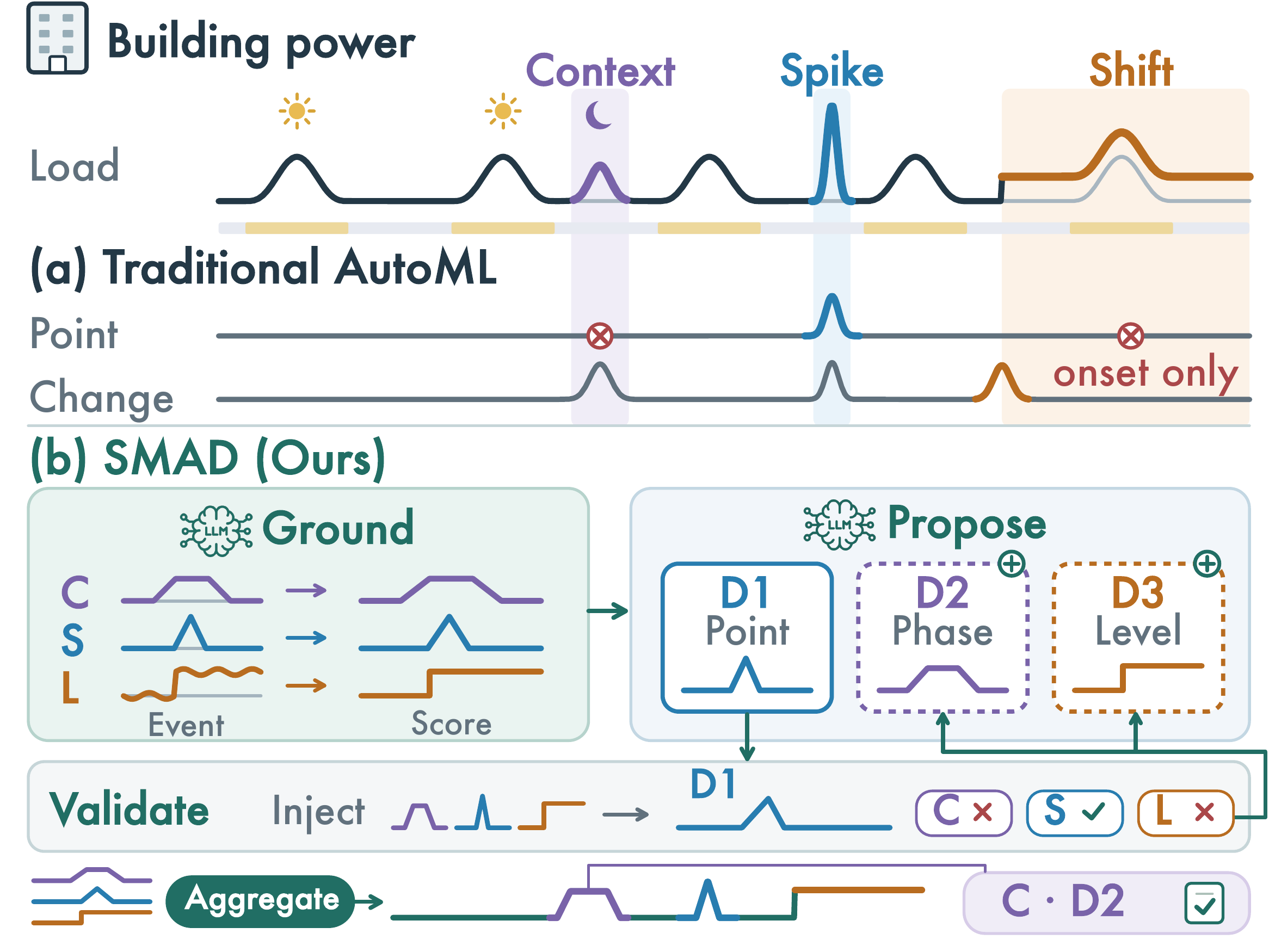}
    \vspace{-0.8cm}
    \caption{Semantics-guided TSAD on an illustrative building-power time series. (a) Traditional AutoML may capture only a subset of anomaly patterns. (b) SMAD uses LLM-based grounding to specify \emph{what to detect} through a semantic contract, which guides planning, label-free validation, and aggregation of detector (D) responses to determine \emph{how to detect}.}
    \vspace{-0.5cm}
    \label{fig:semantic_motivation}
\end{figure}

\noindent \textbf{Motivation.}
This heterogeneity produces a \emph{no-one-size-fits-all} effect: detector rankings vary across datasets and anomaly types~\citep{bahri2022automl,liu2025tsb}. Existing AutoML primarily addresses \emph{how to detect} by selecting or combining detectors using historical performance or label-free proxies, while often leaving \emph{what to detect} implicit. Figure~\ref{fig:semantic_motivation}(a) illustrates this mismatch: nighttime building load can be anomalous despite being normal during daytime. Point scores may capture spikes but miss context and shifts, whereas change scores may capture only shift onsets. More candidates alone cannot resolve this objective mismatch.

We therefore make \emph{what should be detected} explicit through \emph{anomaly semantics}: a structural specification linking event patterns of interest to expected detector-score responses. Grounded in time-series properties and available domain context, a semantic contract links events to expected scores: elevated throughout contextual (C) events, localized peaks for spikes (S), and sustained responses for level shifts (L). In Figure~\ref{fig:semantic_motivation}(b), this contract guides LLM proposals and shared label-free validation through synthetic injections. D1 covers spikes but leaves C/L gaps, prompting phase (D2) and level (D3) specialists. Aggregation combines their responses with detector attribution. Semantics thus makes objectives \emph{inspectable} and directs search toward unmet requirements beyond historical rankings or aggregate proxies.

\noindent \textbf{Challenges.}
The first challenge is an \emph{underspecified target}. Deployment-time TSAD typically lacks anomaly labels and a universally applicable objective function. When multiple anomaly types coexist, their priorities and expected detector-score responses must be explicit and remain stable throughout candidate search.
The second is \emph{label-free validation}. Classical AutoML ranks candidates using held-out labels~\citep{o2023review}, while classification and forecasting often provide accuracy or prediction error as optimization signals~\citep{revin2023automated,alsharef2022review}. Unsupervised TSAD lacks comparably reliable feedback.
The third is \emph{cold start under distribution heterogeneity}. Historical performance can inform detector selection at inference time, but even closely related datasets may provide unreliable guidance under domain or anomaly-morphology shifts.
Finally, automated results often lack \emph{auditability}. When predictions depend on retrieved precedents, planner hypotheses, synthetic tests, and multiple detectors, each alert must be traceable to its contributing semantics, candidates, and supporting evidence.

\noindent \textbf{Proposed approach.}
To address these challenges, we introduce \system, an LLM-based multi-agent framework that makes anomaly intent explicit and grounds detector construction in label-free evidence. A grounding agent derives a fixed contract from measurable time-series properties, defining anomaly types, priorities, expected event and score behaviors, and validation questions. A planner proposes executable pipelines or semantic scorers with explicit targets. Optional benchmark cards from a disjoint meta-training split provide uncertainty-aware priors to mitigate cold start. A shared, candidate-independent protocol evaluates semantic coverage through synthetic anomaly recovery and original-series checks of score morphology, localization, and restraint in low-score regions. Coverage estimates and cross-detector agreement diagnostics guide subsequent proposals toward unmet semantic requirements.
% A grounding agent combines measurable time series properties with optional domain context to construct a frozen semantic contract specifying anomaly types, descriptions, priorities, expected event and score behaviors, and validation questions. Guided by this contract, a planner proposes executable detector pipelines or semantic scorers with declared semantic targets. Optional benchmark cards from a disjoint meta-training split supply uncertainty-aware priors, grounding detector recommendations in historical evidence to mitigate cold start. A shared, candidate-independent validation syllabus assesses synthetic anomaly recovery and checks score morphology, localization, and low-score-region restraint on the original series. These checks estimate semantic coverage, while cross-detector agreement provides diagnostic feedback. Together, they guide subsequent proposals toward unmet semantic requirements.

Moreover, a two-level semantics-aware ensemble combines candidates while preserving timestamp-level semantic and candidate contributions. An append-only evidence bank records decisions and supporting evidence, enabling claim-level auditing and a traceable detection report.
On the univariate and multivariate benchmarks~\citep{liu2024elephant}, \system\ improves VUS-PR~\cite{boniol2025vus} over the strongest AutoML baseline by $11.5\%$ and $9.6\%$, respectively. It also outperforms general coding agents at $2$--$5\times$ lower language-model token cost.

% \noindent \textbf{Contributions.}
Our contributions are fourfold:
\begin{itemize}
[noitemsep,topsep=0pt,parsep=0pt,partopsep=0pt,leftmargin=0.5cm]
\item \textbf{Anomaly intent as a testable objective.} We introduce frozen semantic contracts that link event patterns to expected score responses, making \emph{what to detect} explicit throughout detector proposal, validation, and aggregation.
\item \textbf{Contract-driven label-free refinement.} We develop a shared validation protocol using synthetic anomaly injection and checks of score shape, temporal localization, and low-score-region restraint to assess semantic coverage and guide iterative detector refinement.
\item \textbf{Evidence-grounded and auditable automation.} We combine calibrated benchmark priors, validation-driven semantic ensembling, and claim-level auditing to reduce cold-start risk while tracing alerts to their contributing detectors and supporting evidence.
\item \textbf{A broader TSAD evaluation landscape.} We complement existing benchmarks with an evaluation spanning traditional AutoML, LLM-based methods, and agent-based solutions across univariate and multivariate TSAD.
\end{itemize}

Section~\ref{sec:related} reviews related work, Section~\ref{sec:sphinx} presents the design of \system, Section~\ref{sec:experiments} reports the experimental evaluation, and Section~\ref{sec:conclusion} concludes with the implications of our work.

\section{Related Work}
\label{sec:related}
\vspace{-0.1cm}

\noindent \textbf{Time-Series Anomaly Detection.}
Time-series anomalies are classified as point (global-distribution deviations), contextual (globally ordinary observations violating local dynamics), or collective (anomalous subsequences of individually normal-looking observations)~\citep{paparrizos2025advances}. Detectors are correspondingly diverse~\citep{schmidl2022anomaly,boniol2024dive}: distance-based methods compare observations with reference patterns or nearest neighbors~\citep{breunig2000lof,boniol2021unsupervised}; density- and isolation-based methods identify sparse regions beyond pairwise comparisons~\citep{liu2008isolation,aggarwal2017introduction}; and prediction-based methods score forecasting or reconstruction residuals~\citep{sakurada2014anomaly,munir2018deepant}, increasingly using adapted time-series foundation models~\citep{ansari2024chronos,das2024decoder,liu2024elephant}. Different assumptions produce varying rankings across datasets and anomaly types, with no consistently dominant detector~\citep{schmidl2022anomaly,liu2024time}. \system\ addresses mismatches between detector biases and deployment-specific anomaly semantics, rather than detector availability.

\noindent \textbf{Automated Solutions for TSAD.}
Automated TSAD spans \emph{selection}, \emph{ensembling}, and \emph{generation}~\citep{liu2025tsb}. Selection chooses from a fixed detector pool using meta-learned performance on historical labeled datasets~\citep{zhao2021automatic,sylligardos2023choose,navarro2023meta} or label-free internal criteria~\citep{goswami2023unsupervised}. Ensembling combines candidates for robustness~\citep{aggarwal2015theoretical}; generation synthesizes new scorers from them~\citep{cao2023autood,ye2023uadb}. Three limitations remain. First, single-metric benchmarking fixes an implicit anomaly objective without specifying target behavior or detecting candidate misalignment. Second, meta-learned priors require similarity to historical datasets, while label-free surrogates remain unreliable across scenarios~\citep{liu2025tsb}; neither reconciles historical priors with target time-series evidence. Third, ensembling and generation improve accuracy by executing large pools, increasing costs with pool size without linking decisions to alerts. \system\ makes objectives explicit and combines priors with target evidence for TSAD.

\noindent \textbf{LLMs for Time-Series Anomaly Detection.}
Direct approaches adapt pretrained language-model backbones to temporal tasks~\citep{zhou2023one,liu2026mllmts} or prompt general-purpose models to score textual or plotted series~\citep{xu2026can,park2025delving}. Results are mixed: multimodal prompting is sensitive to serialization, resolution, and normalization and degrades on long or fine-grained time series; effectiveness depends on task formulation; and advantages over conventional temporal models remain contested~\citep{tan2024language}. \system\ combines LLM-based semantic reasoning and pipeline design with deterministic numerical scoring, tying language-model cost to planning rather than time series length.

\noindent \textbf{LLM Agents for Automated Analysis.}
Agents extend individual LLM calls into complete pipelines. AutoML-Agent~\citep{AutoML_Agent} translates natural-language tasks into ML pipelines through role-specialized agents; AD-AGENT~\citep{yang2025ad} builds end-to-end AD pipelines from user requests using existing libraries. General-purpose coding agents, included as baselines, likewise generate detectors for a given time series. These systems follow reflective self-refinement~\citep{madaan2023self,shinn2023reflexion}, requiring feedback -- test suites, execution traces, or validation metrics -- not inherently available in unsupervised TSAD. They derive targets from prompts and refine using supervised metrics or agent judgment, with token costs proportional to trial and error. \system\ grounds refinement in explicit anomaly objectives and label-free validation, providing structured feedback and preserving alert traceability.

\section{\system}
\label{sec:sphinx}

\begin{figure*}[t]
    \centering
    \includegraphics[width=\textwidth]{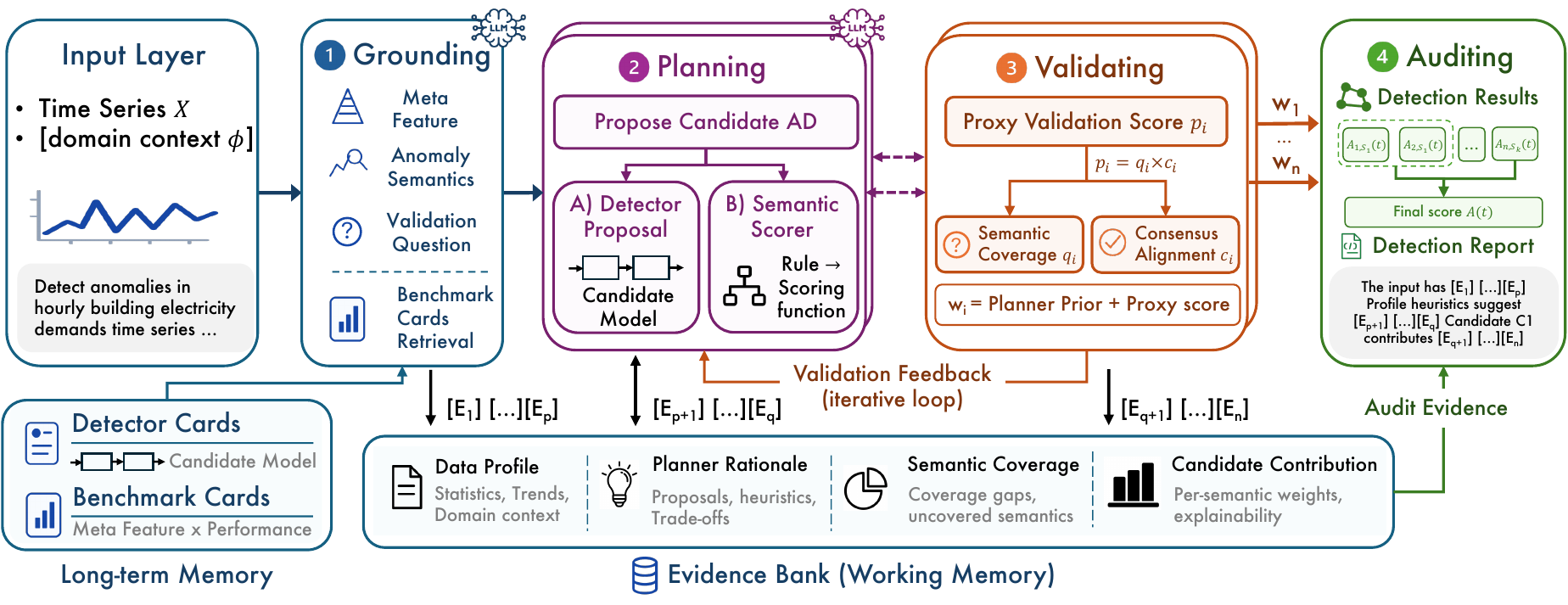}
    \vspace{-0.9cm}
    \caption{Overview of \system. LLMs perform semantic grounding and candidate planning using input evidence and reusable knowledge. Candidate execution, label-free validation, aggregation, and claim auditing are deterministic. Validation feedback guides iterative refinement, while a run-specific evidence bank supports traceable outputs.}
    \vspace{-0.5cm}
    \label{fig:sphinx-architecture}
\end{figure*}

Let $X=(x_1,\ldots,x_T)\in\mathbb{R}^{T\times N}$ be a time series with $N\ge1$ channels. Without anomaly labels, the goal is to produce an anomaly score $A(t)$, $t=\{1,\ldots, T\}$, together with a report. 
As shown in Figure~\ref{fig:sphinx-architecture}, \system\ comprises three components: a \textit{long-term memory} of detector cards and benchmark cards built offline; an \textit{evidence bank} that serves as working memory, recording every decision and its supporting evidence; and a \textit{main workflow} of four core steps. (1)~\textit{Semantic grounding} profiles the input and produce detection objectives (Section~\ref{sec:sphinx-grounding}). (2)~\textit{Planning} proposes candidates guided by these objectives, retrieved memory, and prior feedback (Section~\ref{sec:sphinx-planning}). (3)~\textit{Label-free validation} assesses each candidate and drives refinement (Section~\ref{sec:sphinx-validation}). (4)~\textit{Aggregation and auditing} combines validated scores by anomaly type and checks explanatory claims against the evidence bank (Section~\ref{sec:sphinx-aggregation-auditing}). Details are available in Appendix~\ref{app:implementation}.

\subsection{Semantic Grounding}
\label{sec:sphinx-grounding}

\noindent\textbf{Establishing detection objectives.}
\system\ begins by profiling the input's temporal and distributional structure, including trend, seasonality, noise, missingness, and cross-channel relationships. These measurements, together with optional domain context, support the construction of a \emph{semantic contract}: a bounded set of operational hypotheses about relevant anomalies. 

Formally, the grounding agent outputs a contract
\begin{equation}
\mathcal{C}=\{S_k\}_{k=1}^{K},\ S_k=(\tau_k,d_k,e_k,b_k,\alpha_k,Q_k),
\label{eq:semantic-contract}
\end{equation}
where $\tau_k$ is an anomaly type, $d_k$ a natural-language operational description, $e_k$ the expected event shape, $b_k$ the expected score behavior, $\alpha_k\in[0,1]$ a priority with $\sum_k\alpha_k=1$, and $Q_k$ a set of validation questions. Types are drawn from a fixed vocabulary: \textit{point spikes}, \textit{level shifts}, \textit{contextual anomalies}, \textit{collective anomalies}, and \textit{correlation changes}. Descriptions and priorities cite the profiled measurements that motivate them, and the questions in $Q_k$ probe injection recovery, localization, score shape, or quiet behavior on normal regions. Each entry thus turns a broad anomaly concept into a testable expectation about \emph{both} the event and the score a suitable detector should produce, which guides candidate construction and validation.

The following example illustrates one contract entry for 25-channel SMAP telemetry characterized by weak autocorrelation and a heavy right tail.

\vspace{2pt}
\noindent\fbox{\parbox{\dimexpr\columnwidth-2\fboxsep-2\fboxrule}{\small
% \textbf{Example semantic contract entry} (SMAP telemetry, 25 channels, weak autocorrelation, heavy right tail).\\
$\tau_1$: point spike;\quad $e_1$: point;\quad $\alpha_1=0.34$ (highest).\\
$d_1$: ``Isolated extreme observations [\dots] are anomalous because the value distribution is strongly right-skewed and heavy-tailed while serial dependence is weak.''\\
$b_1$: ``Sharp peaks at isolated extreme samples with rapid return to baseline.''\\
$Q_1$: ``Does the score peak when isolated high-amplitude spikes are injected relative to each channel's local noise scale?''; ``Do high scores remain rare in typical noisy regions without extreme values?''
}}
\vspace{2pt}

The same vocabulary adapts to the input: for a strongly periodic ECG time series, the contract instead ranks contextual (wrong-phase) anomalies first and demotes point spikes to third. Appendix~\ref{app:contract-examples} reproduces both complete contracts.

The contract is frozen before candidate search. This gives all candidates a common target, makes progress measurable through semantic coverage, and helps distinguish failures of the detection objective from failures of the detector. The contract remains an inspectable hypothesis; it does not establish a uniquely correct interpretation of the data.

\noindent\textbf{Using prior knowledge to guide proposals.}
The planner draws on reusable detector cards, benchmark cards constructed from offline data disjoint from the target time series. Detector cards describe available methods, while benchmark cards summarize their behavior on historical time series. Retrieval uses similarity in time series characteristics and the reliability of supporting evidence to form planning priors (card construction and retrieval are detailed in Appendix~\ref{app:benchmark-cards}).

These priors expose both support and uncertainty, helping the planner identify promising candidates while recognizing weak or conflicting precedents. Historical performance influences which candidates are explored, but does not serve as a deployment label or directly determine ensemble weights. Candidate contributions are assessed using evidence from the current run.

\subsection{Planning through Iterative Feedback}
\label{sec:sphinx-planning}

At each iteration, the planner reads the frozen contract, retrieved priors, remaining budget, and validation feedback, and either proposes a catalog detector pipeline, composes a semantic scorer, requests further profiling, or stops. Each proposal $i$ declares its target entries $\mathcal{T}_i\subseteq\mathcal{C}$, a prior $\pi_i\in[0,1]$, the expected score signal, and a rationale citing evidence-bank records, all recorded before execution. Validation (see Section~\ref{sec:sphinx-validation}) maintains a coverage record $\bar q_k=\max_{i:\,S_k\in\mathcal{T}_i}q_{i,k}$ over the candidates validated so far, and the planner targets the highest-priority entry with the largest remaining gap.
% (Figure~\ref{fig:coverage-trajectory}).

\begin{figure}[t]
    \centering
    \includegraphics[width=\columnwidth]{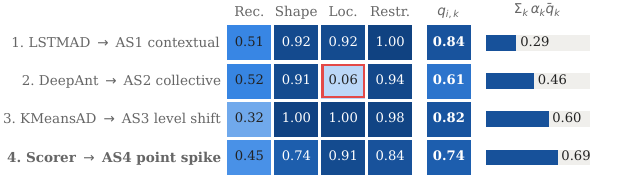}
    \vspace{-0.6cm}
    \caption{Illustration of the planning and refinement step. Each row: candidate and target entry, its validation signals and mean $q_{i,k}$, and resulting coverage $\sum_k\alpha_k\bar q_k$.}
    % The red cell is a failing signal fed back to the planner.}
    \vspace{-0.6cm}
    \label{fig:coverage-trajectory}
\end{figure}

Figure~\ref{fig:coverage-trajectory} traces this loop on UCR \#79, where AS$k$ (anomaly semantic $k$) denotes contract entry $S_k$ in priority order. LSTMAD covers AS1 and lifts coverage to $0.29$; DeepAnt only partially covers AS2 (localization $0.06$), reaching $0.46$; KMeansAD then targets AS3, reaching $0.60$.

\noindent\textbf{Semantic scorers.}
Some entries match the inductive bias of no catalog detector. For these, the planner composes a \emph{semantic scorer}, translating the entry's description $d_k$ and expected score behavior $b_k$ into a weighted program of deterministic, label-free operators such as local residuals, differences, or changepoint contrasts (Appendix~\ref{app:semantic-scorers}). For the remaining point-spike entry AS4 in Figure~\ref{fig:coverage-trajectory}, the planner argues that on a smooth waveform a narrow impulse is best exposed by local residuals and curvature, and composes $0.35\,\text{median residual}_{11}+0.30\,|\Delta^2x|+0.20\,|\Delta x|+0.15\,\text{robust }z\text{-score}$, where $\Delta$ is the first difference and the subscript is the window length. It raises coverage to $0.69$. A gate admits a scorer only after a catalog detector has run and an entry remains weakly covered, and the scorer is validated and aggregated like any detector.  Appendix~\ref{app:scorer-cases} illustrates how a scorer complements catalog detectors.

\subsection{Label-Free Validation and Self-Refinement}
\label{sec:sphinx-validation}

\noindent\textbf{A shared evaluation protocol.}
Before any candidate runs, \system\ constructs a shared suite of synthetic interventions from the contract, prioritizing important semantics while limiting redundancy within a fixed budget. If the contract cannot be mapped to supported interventions, \system\ falls back to data-derived characteristics, domain priors, or a default suite. All candidates are evaluated on this suite, preventing them from choosing their own tests and ensuring comparable validation feedback.

\noindent\textbf{Semantic coverage.}
For each entry $S_k\in\mathcal{T}_i$ targeted by candidate $i$, validation computes four signals in $[0,1]$ with the same fixed criteria for every candidate (details in Appendix~\ref{app:validation-weights}):
\begin{itemize}[noitemsep,topsep=2pt,parsep=0pt,partopsep=0pt,leftmargin=0.4cm]
    \item \emph{Intervention recovery} $r_{i,k}$: whether injected anomalies receive high scores.
    \item \emph{Score shape} $h_{i,k}$: whether the score profile matches the expected behavior $b_k$, such as isolated peaks or sustained elevation.
    \item \emph{Localization} $\ell_{i,k}$: whether high-scoring regions have the temporal extent implied by the event shape $e_k$.
    \item \emph{Restraint} $z_{i,k}$: whether scores stay low in the candidate's own low-score region.
\end{itemize}
% Recovery is measured on the intervention runs and the other three signals on the original time series (exact rules in Appendix~\ref{app:validation-weights}). 
The semantic coverage of candidate $i$ is defined as 
\begin{equation}
\begin{aligned}
q_{i,k}&=\tfrac14\bigl(r_{i,k}+h_{i,k}+\ell_{i,k}+z_{i,k}\bigr),\\
q_i&=\frac{1}{|\mathcal{T}_i|}\sum_{S_k\in\mathcal{T}_i}q_{i,k}.
\end{aligned}
\label{eq:semantic-coverage}
\end{equation}
Coverage measures agreement with the declared semantics, not detection accuracy; it provides a comparable, label-free basis for refinement.

\noindent\textbf{Refinement and consensus alignment.}
Each $q_{i,k}$ updates the planner's coverage record $\bar q_k$ (Section~\ref{sec:sphinx-planning}), and the individual signals fed back with it separate poor recovery, mismatched shape, inaccurate localization, and execution failure. In Figure~\ref{fig:coverage-trajectory}, DeepAnt's localization of $0.06$ leaves the collective entry only partially covered, so later proposals, including a semantic scorer, target the remaining gaps and raise priority-weighted coverage from $0.29$ to $0.69$.
To complement these per-candidate checks, \system\ computes a \emph{consensus alignment} $c_i\in[f_{\min},1]$ from each candidate's mean Spearman rank agreement with the rest of the pool: nonnegative agreement gives $c_i=1$, negative agreement lowers it linearly, and rank-agreement outliers receive $f_{\min}=0.4$ (Eq.~\eqref{eq:consensus-alignment}). The floor limits penalties for disagreement that may reflect complementary anomaly sensitivity. Alignment is fed back to the planner and recomputed over the full pool before aggregation.

\noindent\textbf{Ensemble weights.}
The proxy validation score $p_i=q_i\,c_i$ rewards candidates that satisfy their declared semantics and remain consistent with the pool. It is blended with the planner prior $\pi_i\in[0,1]$, which the planner fixes before execution:
\begin{equation}
\tilde w_i{=}\frac{n_i\,p_i{+}\pi_i}{n_i{+}1},\
w_i{=}(1{-}\eta)\frac{\tilde w_i}{\sum_{j\in\mathcal U}\tilde w_j}{+}\frac{\eta}{|\mathcal U|},
\label{eq:candidate-weight}
\end{equation}
where $n_i$ counts the validation evidence behind $p_i$ (intervention specifications $\times$ validation-frame points), and $\mathcal U$ is the set of numerically usable candidates. The prior thus matters only when evidence is thin; in the reported runs its share is below $0.2\%$. The blend toward uniform weights ($\eta=0.15$) prevents collapse onto a single noisy proxy.

\subsection{Evidence-Calibrated Aggregation and Auditing}
\label{sec:sphinx-aggregation-auditing}
\vspace{-0.1cm}

\noindent\textbf{Semantic-aware ensemble.}
Aggregation is performed in two stages: candidates are first averaged within each semantic group, and the groups are then combined. Let $\tilde A_i(t)$ be the robust $z$-score of candidate $i$'s score sequence $A_i(t)$, and let group $G_k=\{i\in\mathcal U: S_k\in\mathcal{T}_i\}$ contain the usable candidates that target $S_k$. A candidate joins the group of each of its targets and splits its weight $w_i$ (Eq.~\eqref{eq:candidate-weight}) evenly among them, $\omega_{i,k}=w_i/|\mathcal{T}_i|$, so it gains no extra influence by declaring more targets. Within each group, the semantic score is the weighted mean
\begin{equation}
A_{S_k}(t)=\zeta\!\left(\frac{\sum_{i\in G_k}\omega_{i,k}\,\tilde A_i(t)}{\sum_{i\in G_k}\omega_{i,k}}\right),
\label{eq:within-group}
\end{equation}
where $\zeta(\cdot)$ standardizes the group stream so that semantics with different numbers of candidates or score ranges are on a common scale. Across groups, the final score takes the strongest semantic response at each timestamp:
\begin{equation}
A(t)=\max_{k}\,A_{S_k}(t).
\label{eq:semantic-ensemble}
\end{equation}
Averaging within a group pools detectors that look for the same anomaly pattern and reduces the noise of any single detector. Taking the maximum across groups prevents an anomaly that only one semantic captures, such as a brief spike among level-shift detectors, from being diluted by unrelated semantics. 
% The maximizing group $k^\ast(t)=\arg\max_k A_{S_k}(t)$ identifies which semantic drives each alert, and the weights $\omega_{i,k^\ast}$ attribute it to individual candidates within that group.

\noindent\textbf{Evidence-based reporting.}
Throughout the run, the evidence bank serves as working memory: it records input profiles, grounding decisions, retrieved knowledge, proposal rationales, validation results, and aggregation records, each with an identifier, stage, and source, and retains every candidate and semantic score stream. 

The report is assembled directly from these records without an additional LLM call. This keeps explanations tied to the decisions made during the run, makes them reproducible from the same evidence, and limits the introduction of unsupported claims during reporting. It summarizes candidate rationales, validation results, and ensemble weights. Each flagged timestamp is attributed to the semantic group $k^\ast(t)=\arg\max_k A_{S_k}(t)$ and its contributing candidates. Sentence-level explanatory claims cite evidence identifiers. A deterministic auditor checks these references and approximate numerical consistency, classifying claims as \textit{supported}, \textit{partially supported}, \textit{unsupported}, or \textit{contradicted}. Unsupported and contradicted claims are removed; partially supported claims are retained with a marker. An example report is in Appendix~\ref{app:report-example}.
% This audit establishes consistency with recorded evidence, with an example report available in Appendix~\ref{app:report-example}.

\section{Experiments}
\label{sec:experiments}
\vspace{-0.2cm}

In this section, we describe the experimental setup (Section~\ref{sec:experimental-setup}) followed by the results (Section~\ref{sec:experimental-results}).

% \qinghua{TODO: clarify benchmark card use, exp using the same set of candidates thus removing autotsad, maybe add a column in tab 1}

\subsection{Experimental Setup}
\label{sec:experimental-setup}

\noindent \textbf{Datasets.}
We follow TSB-AutoAD's protocol~\citep{liu2025tsb} on TSB-AD collections~\citep{liu2024elephant}, restricted by licensing to twelve sources (UCR, MGAB, TODS, Daphnet, OPP, SMD, SMAP, GECCO, MITDB, SVDB, LTDB, CATSv2) spanning six domains and point, contextual, and collective anomalies. We retain its domain-aware split (Table~\ref{tab:benchmark_datasets}) into meta-training (262 univariate, 83 multivariate series) and evaluation (115 univariate, 31 multivariate series), with no shared time series. For fairness, meta-training is the sole historical data available to all methods, used to build \system{}'s benchmark cards and train the meta-learning selectors. The evaluation set serves only to assess automated solutions; no deployable method accesses its labels. Appendix~\ref{app:dataset-descriptions} explains the restriction and describes the sources.

\noindent \textbf{Candidate models.}
The candidate model set defines which anomaly detection approaches are available, and every method that selects, ensembles, or generates from a pool uses the same set, so all methods are compared on one testbed. It contains $15$ TSB-AD detectors (Table~\ref{tab:candidate}) with default hyperparameters.
% : eight statistical (MCD, OCSVM, LOF, KNN, KMeansAD, IForest, HBOS, PCA) and seven neural (AutoEncoder, LSTMAD, DeepAnt, Donut, OmniAnomaly, USAD, TranAD). 
This set is exactly what \system's detector cards describe. \system{} and the coding agents receive the same neutral catalog (names, import paths, defaults) without quality guidance.

\noindent \textbf{Baselines and metrics.}
To rigorously evaluate the proposed \system, we consider a comprehensive suite of baselines spanning four categories:
\begin{itemize}[noitemsep,topsep=2pt,parsep=0pt,partopsep=0pt,leftmargin=0.4cm]
    \item \emph{Baseline solutions}: Global Best, Supervised Selection, and Random Ensemble, which aggregates randomly selected scores~\citep{liu2025tsb};
    \item \emph{AutoML}: the top performing baseline from TSB-AutoAD AvgEns~\citep{aggarwal2015theoretical}, which averages all candidate scores; the meta-learning selectors MetaOD~\citep{zhao2021automatic} and MSAD~\citep{sylligardos2023choose}; the surrogate metrics based model selector TSADAMS~\citep{goswami2023unsupervised}; and the automated system AutoOD~\citep{cao2023autood};
    \item \emph{LLM-based}: LLMAD, MLLMAD~\citep{park2025delving}, and MLLM4TS~\citep{liu2026mllmts}, which use LLMs as standalone anomaly detectors through prompting or fine-tuning;
    \item \emph{Coding agents}: Claude Code (Opus 4.8)~\citep{anthropic_claude_2026}, Codex (GPT-5.5)~\citep{openai_codex_2026}, and AD-AGENT~\citep{yang2025ad}.
\end{itemize}
All baselines are reproduced and evaluated under the same setup as \system{}, with the same meta-training/evaluation split and the same candidate model set. Unless otherwise specified, AD-AGENT and \system\ also use GPT-5.5 to ensure a consistent language-model backbone for comparison. For fair comparison, methods that call an LLM have a cap of $600$\,s per time series; if a run exceeds it, the system returns the best solution found so far.
Following prior work, we evaluate scoring quality using AUC-PR and VUS-PR~\citep{paparrizos2022volume,boniol2025vus}. Appendix~\ref{app:setup-details} provides detailed descriptions of the baselines.

\vspace{-0.1cm}
\subsection{Experimental Results}
\label{sec:experimental-results}
\vspace{-0.1cm}

% We first evaluate detection performance and computational efficiency, then examine how iterative refinement, semantic aggregation, and optional domain context contribute to \system{}.

% Auto-generated by draft/exp_results/build_exp_results.py.
% Regenerate: python draft/exp_results/build_exp_results.py
\begin{table}[t]
\centering
\caption{Performance overview of \system{} and baselines. Values are per-series means; higher is better. Families: \faTag~Baseline Solution, \faCogs~AutoML, \faCommentsO~LLM-based, \faCode~coding agent, \faStar~ours. \textbf{Hist.}: has access to meta-training datasets. \textbf{Run-All}: must execute every candidate detector on the target series. \textbf{Bold} marks the best and \underline{underlined} the second best. Additional per-dataset evaluation results are available in Table~\ref{tab:per-dataset}.}
\vspace{-0.3cm}
\label{tab:main-comparison}
\providecommand{\cmark}{\ding{51}}\providecommand{\xmark}{\ding{55}}
% \small
% \setlength{\tabcolsep}{4pt}
\resizebox{\columnwidth}{!}{%
\begin{tabular}{clcccccc}
\toprule
 & \multirow{2}{*}{\textbf{Method}} & \multirow{2}{*}{\textbf{Hist.}} & \multirow{2}{*}{\textbf{Run-All}} & \multicolumn{2}{c}{\textbf{Univariate}} & \multicolumn{2}{c}{\textbf{Multivariate}} \\
\cmidrule(lr){5-6} \cmidrule(lr){7-8}
& & & & \textbf{AUC-PR} & \textbf{VUS-PR} & \textbf{AUC-PR} & \textbf{VUS-PR} \\
\midrule
\multirow{3}{*}{\faTag} & Global Best & \cmark & \xmark & 0.3534 & 0.3712 & 0.2506 & 0.2510 \\
 & Supervised Selection & \cmark & \xmark & 0.4502 & 0.4826 & 0.3003 & 0.3087 \\
 & Random Ensemble & \xmark & \xmark & 0.2146 & 0.2519 & 0.1684 & 0.1750 \\
\midrule
\multirow{5}{*}{\faCogs} & AvgEns & \xmark & \cmark & \underline{0.4206} & \underline{0.4673} & 0.2663 & 0.2906 \\
 & MetaOD & \cmark & \xmark & 0.3000 & 0.3479 & 0.2495 & 0.2632 \\
 & MSAD & \cmark & \xmark & 0.2993 & 0.3609 & \underline{0.2998} & \underline{0.3133} \\
 & TSADAMS & \xmark & \cmark & 0.1879 & 0.2278 & 0.1337 & 0.1319 \\
 & AutoOD & \xmark & \cmark & 0.1352 & 0.1687 & 0.1741 & 0.1693 \\
\midrule
\multirow{3}{*}{\faCommentsO} & LLMAD & \xmark & \xmark & 0.1205 & 0.1428 & -- & -- \\
 & MLLMAD & \xmark & \xmark & 0.1228 & 0.1412 & -- & -- \\
 & MLLM4TS & \xmark & \xmark & -- & -- & 0.2228 & 0.2554 \\
\midrule
\multirow{3}{*}{\faCode} & Claude Code & \xmark & \xmark & 0.3110 & 0.3575 & 0.2168 & 0.2358 \\
 & Codex & \xmark & \xmark & 0.3121 & 0.3343 & 0.1788 & 0.1888 \\
 & AD-AGENT & \xmark & \xmark & 0.2852 & 0.3305 & 0.2205 & 0.2298 \\
\midrule
\multirow{2}{*}{\faStar} & \system{} (w/o bench) & \xmark & \xmark & 0.3779 & 0.4093 & 0.2721 & 0.2651 \\
 & \system{} & \cmark & \xmark & \textbf{0.4697} & \textbf{0.5212} & \textbf{0.3034} & \textbf{0.3432} \\
\bottomrule
\end{tabular}%
}
\vspace{-0.3cm}
\end{table}

\noindent \textbf{Main comparison.}
\system{} achieves the strongest overall performance, ranking first on both metrics for univariate and multivariate time series (Table~\ref{tab:main-comparison}). Its gains over AutoML methods and general-purpose coding agents highlight the effectiveness of the proposed workflow, while substantial improvements over LLMAD and MLLMAD support using LLMs to guide executable detection pipelines. Even without benchmark cards, \system{} outperforms Global Best, which uses labeled information. Benchmark cards further improve performance by steering proposals toward detectors that performed well on related time series, allowing \system{} to surpass other supervised selection and AutoML baselines given the same historical information. \system{} also outperforms the previously leading AvgEns baseline while avoiding its requirement to execute all candidate models, offering a more computationally efficient alternative.

\begin{figure}[t]
\centering
\includegraphics[width=\columnwidth]{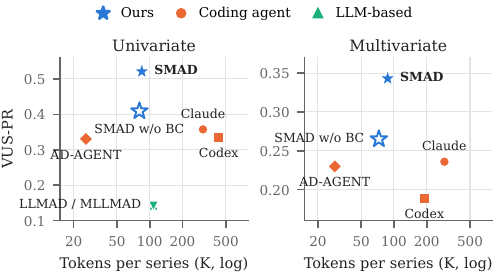}
\vspace{-0.7cm}
\caption{Token cost against accuracy. Each point is a method that calls a hosted LLM, placed at its mean tokens per evaluation time series (log scale) and its mean VUS-PR; the upper left is better.}
\vspace{-0.6cm}
\label{fig:token-cost}
\end{figure}

\noindent \textbf{Token cost.} 
Across both splits, \system{} outperforms Claude Code and Codex while consuming on average $3.6\times$ (up to $5.1\times$) fewer tokens, yielding a favorable accuracy--cost trade-off (Figure~\ref{fig:token-cost}). Reasoning focuses on grounding and planning, while deterministic execution and validation provide structured feedback. The fixed contract and shared intervention syllabus focus exploration and limit context growth. \system{} also uses fewer tokens than LLMAD and MLLMAD, which repeatedly submit time-series windows to the model. AD-AGENT is cheaper, but its single-detector approach yields substantially lower detection accuracy. These comparisons show that \system{} uses its additional planning budget productively.

\begin{figure}[t]
\centering
\includegraphics[width=\columnwidth]{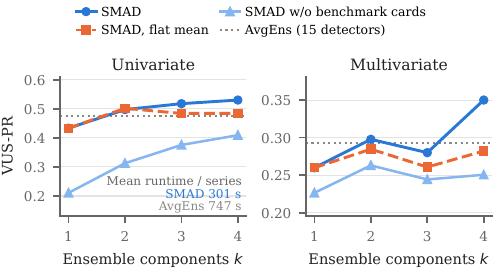}
\vspace{-0.8cm}
\caption{Effect of ensemble size $k$. \system{} uses semantic aggregation (Eqs.~\ref{eq:within-group}--\ref{eq:semantic-ensemble}); the flat baseline takes a weighted mean of the same candidates.}
\vspace{-0.5cm}
\label{fig:ensemble-size}
\end{figure}

\noindent \textbf{Number of ensemble components.} 
\system{} builds a compact, effective ensemble through targeted refinement. We replay final aggregation using the first $k$ candidates in proposal order with renormalized weights. On the univariate split, only two candidates are needed to surpass AvgEns, which averages all $15$ detectors, and each subsequent candidate further improves performance (Figure~\ref{fig:ensemble-size}). With four candidates, \system{} achieves stronger detection performance while requiring approximately $2.5\times$ less mean runtime than AvgEns ($301$\,s vs.\ $747$\,s; per-method runtimes in Appendix~\ref{app:runtime}). The multivariate ensemble also achieves its best replayed performance at $k=4$.

With the same candidate pool, semantic aggregation achieves higher accuracy than a flat weighted mean, preserving complementary responses that can be attenuated when scores are averaged across anomaly types. The gains thus extend beyond candidate selection to combination by detection objective. Benchmark memory and iterative refinement are complementary: cards substantially improve initial proposals, while refinement narrows the performance gap when cards are absent. These findings support using prior experience for a strong starting point and deployment-time feedback to improve candidate selection.

\begin{figure}[t]
    \centering
    \includegraphics[width=\columnwidth]{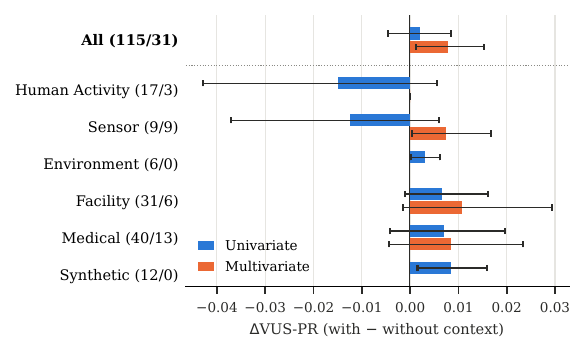}
    \vspace{-0.8cm}
    \caption{Effect of domain context on \system{} by domain: paired mean VUS-PR change (with $-$ without context) with 95\% bootstrap intervals. Parentheses give univariate/multivariate series counts. Context augments the same frozen no-context ensemble.}
    \vspace{-0.6cm}
    \label{fig:domain-context}
\end{figure}

\noindent \textbf{Effect of domain context.} 
\system{} achieves its main results without domain context. We assess its added value by replaying each time series with a brief dataset description of the measured signal and channel roles (see details in Appendix~\ref{app:domain-context}). One additional LLM call may propose a domain-informed scorer, admitted to the frozen ensemble only if it passes a label-free benefit gate. Context helps when descriptions specify actionable temporal structure or channel relationships, as shown in Figure~\ref{fig:domain-context}. 
For multivariate Sensor time series, descriptions distinguishing command channels from measured targets improve VUS-PR. Domain context such as beat alignment likewise yields mean VUS-PR gains across $12$ univariate ECG time series. Univariate Human Activity and Sensor results emphasize description quality: in these single-channel settings, descriptions primarily reference unavailable information, and generic variance scorers can respond to normal activity changes. 
Together, these findings favor operational descriptions with precise channel roles and domain-specific temporal references that guide scorer construction toward meaningful anomaly patterns, strengthening an already effective pipeline.

\noindent \textbf{Case study.}
We trace one time series through three code-generating systems. \system{} first defines its targets: contextual and collective deviations from the dominant cycle. Its initial KNN candidate only partly covers the top contextual entry ($q=0.344$), prompting the planner to target the uncovered collective entry \emph{``with a density-ratio detector rather than nearest-neighbor distance magnitude''}. The resulting LOF achieves the highest coverage ($q=0.851$), while later validation flags KMeansAD as unstable under perturbation. The ensemble achieves $0.684$ VUS-PR using $88$K tokens. Claude Code instead stops at one generic detector ($0.006$ VUS-PR), while Codex rank-sums $19$ heterogeneous signals but discards the two effective detectors ($0.010$ VUS-PR, $673$K tokens). Appendix~\ref{app:trajectory} provides the detailed comparison.

\vspace{-0.1cm}
\section{Conclusion}
\label{sec:conclusion}
\vspace{-0.2cm}

\system\ advances automated TSAD by translating explicit anomaly intent into executable detection pipelines. A frozen semantic contract focuses planning, benchmark memory informs candidate proposals, and iterative label-free validation guides refinement. Semantic aggregation preserves complementary candidate responses, while recorded evidence makes the resulting scores and explanations inspectable. \system{} achieves the strongest overall performance among a comprehensive suite of AutoML, LLM-based, and agent-based baselines. These findings demonstrate the value of structured agentic workflows in combining detection accuracy, efficient LLM usage, and traceable outputs. Future work could explore agentic reinforcement learning to learn more effective planning and refinement policies from accumulated validation feedback.

\clearpage
\newpage

\section*{Limitations}

\system{} focuses on workflow construction and multi-agent orchestration through a predefined sequence of grounding, planning, validation, and aggregation, with explicit agent roles and output formats. The language models operate through prompting without task-specific training, and validation feedback guides refinement within each episode. This design leaves opportunities to explore alternative workflow structures and extend planner learning across episodes.

A promising direction is to incorporate agentic reinforcement learning. The label-free validation signals already recorded by \system{}, including semantic coverage and stability under perturbation, could serve as rewards for training planning and refinement policies. Such training could help the system learn which proposals satisfy specific anomaly semantics, complementing the base model's priors with task-specific experience.


\bibliography{ref}

\clearpage
\newpage

\appendix

\section*{Supplementary Material for \system}

The appendix is organized in two parts. Appendix~\ref{app:impl-details} gives the implementation details needed to reproduce \system: the fixed evaluation instantiation and validation weights (Appendix~\ref{app:implementation}), the construction and retrieval of benchmark cards (Appendix~\ref{app:benchmark-cards}), the semantic scorers (Appendix~\ref{app:semantic-scorers}), and the experimental setup, datasets, and candidate detectors (Appendices~\ref{app:setup-details} and~\ref{app:dataset-descriptions}). Appendix~\ref{app:extra-results} reports additional results, including per-dataset scores (Appendix~\ref{app:per-dataset}), consistency across domains (Appendix~\ref{app:domain-consistency}), token and runtime costs (Appendices~\ref{app:token-accounting} and~\ref{app:runtime}), and qualitative artifacts from the reported runs: semantic scorer case studies, semantic contract examples, an end-to-end run trajectory, a complete detection report, and domain-context examples (Appendices~\ref{app:scorer-cases}--\ref{app:domain-context}).

\section{Implementation Details}
\label{app:impl-details}

This appendix gives the fixed configuration, the benchmark-card memory, the semantic scorers, and the experimental setup and datasets.

\subsection{Fixed Evaluation Instantiation}
\label{app:implementation}

The method is not tied to the numerical choices below: budgets may be changed
to match deployment cost, and all validation and aggregation constants are
ordinary configuration rather than learned parameters.  For the reported
evaluation, however, we use one fixed configuration for every univariate and
multivariate time series.  No global configuration value is selected separately for
an individual dataset or target time series; only the planner's intended
time series-conditioned decisions vary between episodes.
Table~\ref{tab:fixed-method-config} records the settings that shape the
method's behavior, together with the rationale for each choice.

\begin{table*}[t]
\centering
\caption{Fixed evaluation instantiation used for the reported runs. These values
define one reproducible operating point rather than requirements of the
framework.}
\label{tab:fixed-method-config}
\small
\begin{tabular}{p{0.18\textwidth}p{0.31\textwidth}p{0.43\textwidth}}
\toprule
\textbf{Component} & \textbf{Fixed setting} & \textbf{Role and rationale} \\
\midrule
Semantic contract
& At most five semantics; four equally weighted validation checks
& Bounds the grounding schema and treats the four label-free checks symmetrically. \\

Planning budget
& Four candidate proposals; 600-second wall-clock cap per episode
& Provides the same bounded planning and execution budget to every time series; the budget changes cost, not the definition of a valid candidate. \\

Benchmark retrieval
& Exact global scan; pool size 30; eight retained cards; profile/reliability coefficients $(1,0.2)$
& Makes time-series structure the primary retrieval signal while record reliability acts as secondary evidence. \\

Semantic-scorer gate
& At least one model candidate; coverage, model-support, benchmark-confidence, and redundancy cutoffs $(0.5,1,0.5,2)$
& Implements a conservative model-first policy: a deterministic scorer is considered only when current evidence indicates an uncovered or weakly supported semantic. \\

Validation syllabus
& At most three specifications; synthetic anomaly rate $0.03$; magnitude scale $0.25$; perturbation-noise scale $0.02$
& Bounds validation cost and uses one sparse, nontrivial intervention battery and one small stability perturbation shared by all candidates in an episode. \\

Validation frame
& Full time series through 2,048 points; otherwise stride-10 sampling; 10,000-point cap with informative contiguous windows
& Bounds repeated validation cost while retaining global temporal coverage; informative windows are used only for very long time series. \\

Ensemble guardrails
& Consensus floor $f_{\min}=0.4$; uniform blend $\eta=0.15$
& Prevents consensus from deleting a potentially complementary stream, and prevents weight collapse onto one candidate. \\

Score normalization
& Candidate robust $z$-score; semantic-stream standard $z$-score
& Places heterogeneous detector and semantic scores on comparable scales. \\
\bottomrule
\end{tabular}
\end{table*}

For completeness, the recovery signal $r_{i,k}$ in
Eq.~\eqref{eq:semantic-coverage} is the mean of the available values
$\operatorname{AP}_i$, $0.8R_i$, and $\operatorname{sig}(3g_i)$, where
$\operatorname{AP}_i$ is the average precision on the injected anomalies,
$R_i$ the recall at the number of injected points, $g_i$ the standardized
gap between the mean scores of injected and normal points, and $\operatorname{sig}$ the
logistic function.  The low-score region of the restraint signal $z_{i,k}$ lies
below the median score.  The shape checks use
the upper $1\%$ set for point isolation, the upper quartile for level-shift
continuity, and the upper $10\%$ set for contextual spread.  Localization uses
the larger of the upper-$5\%$ cutoff and half the maximum normalized score.
Unknown event shapes receive the neutral value $0.5$; constant streams receive
zero, and a low-score region with fewer than two eligible points receives one.
Global upper-tail agreement uses the top $1\%$ indices.  These order-statistic
rules are scale-free and are applied unchanged to every candidate and dataset.

\paragraph{Consensus and weighting formulas.}
\label{app:validation-weights}
Let $\mathcal U$ be the numerically usable candidates. For consensus alignment, let $\rho_{ij}$ be the Spearman correlation between the score sequences of candidates $i$ and $j$ on the original time series, and
\begin{equation*}
\bar\rho_i=\frac{1}{|\mathcal U|-1}\sum_{j\neq i}\frac{1+\rho_{ij}}{2}.
\end{equation*}
The outlier set $\mathcal O$ holds the candidates with $\bar\rho_i<\mu_\rho-\max(0.1,\sigma_\rho)$, where $\mu_\rho$ and $\sigma_\rho$ are the mean and standard deviation of $\{\bar\rho_j\}$. Then
\begin{equation}
c_i=
\begin{cases}
f_{\min}, & i\in\mathcal O,\\
f_{\min}+(1-f_{\min})\min(1,2\bar\rho_i), & i\notin\mathcal O,
\end{cases}
\label{eq:consensus-alignment}
\end{equation}
and $c_i=1$ when fewer than two usable candidates exist. 
% In Eq.~\eqref{eq:candidate-weight}, the evidence count $n_i$ is the number of intervention specifications times validation points ($n_i\ge615$ in the reported runs), so the prior share $1/(n_i+1)$ is below $0.2\%$.

Algorithm~\ref{alg:sphinx} summarizes the main end-to-end execution path.  The pseudocode separates the model-mediated decisions--grounding and planning--from the deterministic execution, validation, aggregation, and audit operations.

\begin{algorithm}[t]
\caption{Main \system{} execution path}
\label{alg:sphinx}
\small
\begin{algorithmic}[1]
\Require Time series $X$, optional context $\phi$, detector cards $\mathcal D$, benchmark cards $\mathcal B$, fixed configuration $\Theta$
\State $F \gets \Call{Profile}{X}$
\State $\mathcal C \gets \Call{Ground}{F,\phi}$ \Comment{freeze the contract, Eq.~\eqref{eq:semantic-contract}}
\State $\mathcal R,\Pi \gets \Call{Retrieve}{F,\mathcal C,\mathcal B}$
\State $\mathcal I \gets \Call{BuildSyllabus}{F,\mathcal C,\phi,\Theta}$ \Comment{shared interventions}
\State $\mathcal P\gets[\,]$; $\mathcal A\gets[\,]$; $\mathcal V\gets[\,]$; $\bar{\mathbf q}\gets\mathbf 0$
\While{proposal and runtime budgets remain}
    \State $a\gets\Call{Plan}{F,\mathcal C,\mathcal R,\Pi,\bar{\mathbf q},\mathcal V}$
    \If{$a$ requests more profiling}
        \State $F\gets\Call{ActiveProfile}{X,a}$; \textbf{continue}
    \ElsIf{$a$ stops}
        \State \textbf{break}
    \EndIf
    \State $P_i\gets\Call{CompileSafeCandidate}{a,\mathcal D}$
    \State $A_i\gets\Call{Execute}{P_i,X}$
    \State $(E_i,\mathbf q_i)\gets\Call{Validate}{P_i,A_i,\mathcal I,\mathcal C}$ \Comment{Eq.~\eqref{eq:semantic-coverage}}
    \State append $(P_i,A_i,E_i)$ to $(\mathcal P,\mathcal A,\mathcal V)$
    \State $\bar{\mathbf q}\gets\max(\bar{\mathbf q},\mathbf q_i)$ \Comment{elementwise over declared targets}
    \State append global agreement diagnostics to the next planner state
\EndWhile
\State $\mathcal U\gets\Call{NumericallyUsable}{\mathcal P,\mathcal A}$
\ForAll{$i\in\mathcal U$}
    \State $c_i\gets\Call{ConsensusAlignment}{A_i,\mathcal A_{\mathcal U}}$
\EndFor
\State $\mathbf w\gets\Call{GuardedWeights}{\{(q_ic_i,\pi_i,n_i):i\in\mathcal U\}}$ \Comment{Eq.~\eqref{eq:candidate-weight}}
\ForAll{semantic groups $k$}
    \State $A_{S_k}\gets\Call{SemanticAggregate}{\mathcal U,\mathbf w,k}$
\EndFor
\State $A(t)\gets\max_k A_{S_k}(t)$
\State \Return $A$ and the audited report
\end{algorithmic}
\end{algorithm}

\subsection{Benchmark Card Construction and Retrieval}
\label{app:benchmark-cards}

Benchmark cards are \system{}'s memory of how the candidate detectors behaved on previously seen time series. They are built once, offline, from the TSB-AutoAD~\citep{liu2025tsb} meta-training split, which shares no time series with the evaluation split. Building them uses training labels only to score the candidates on each training time series; at deployment no label of the target time series is read.

\noindent\textbf{Card contents.} Each card describes one training time series $s$ and stores four fields.
\begin{itemize}[noitemsep,topsep=2pt,parsep=0pt,partopsep=0pt,leftmargin=0.4cm]
    \item \emph{Anchor features} $\mathbf{a}_s$, the retrieval key. For a univariate time series these are the 22 catch22 features~\citep{lubba2019catch22}. For a multivariate time series, catch22 is computed per channel and each feature is summarized across channels by its minimum, first quartile, mean, third quartile, and maximum, giving $110$ values regardless of the channel count.
    \item The \emph{profile} produced by the same deterministic profiler used online (trend, seasonality, noise, autocorrelation, distribution shape, cross-channel correlation), together with the anomaly types the grounding agent suspects for the time series. These types come from the same LLM call used in semantic grounding, so cards and queries share one anomaly-type vocabulary. This is a one-time offline cost; retrieval adds no LLM call at deployment.
    \item \emph{Detector results}: the VUS-PR $v_{s,m}$ of every catalog detector $m\in\mathcal D$ on $s$, from the same TSB-AD implementations and default hyperparameters that \system{} executes, together with the resulting ranking of the candidates.
    \item Metadata: time series length, number of labeled anomalies, and the source dataset.
\end{itemize}

\noindent\textbf{Retrieval.} For the target time series $X$, \system{} computes the same anchor features $\mathbf{a}_X$ and standardizes query and cards with statistics from the training split, denoted $\zeta(\cdot)$. Features are aligned by name, so the comparison is not affected by feature order. With $\delta_s=\lVert \zeta(\mathbf{a}_X)-\zeta(\mathbf{a}_s)\rVert_2$, profile similarity is $\mathrm{sim}_s=1/(1+\delta_s)$. An exact scan keeps the $30$ closest cards. These are reranked by $\mathrm{sim}_s+0.2\,\lambda_s$, where the reliability
\begin{align*}
\lambda_s ={}& 0.6\,v^{\max}_s + 0.2\,\min\!\left(1,\tfrac{|\mathcal D_s|}{12}\right)\\
&+ 0.2\,\frac{v^{\max}_s}{1+\mathrm{sd}_m(v_{s,m})}
\end{align*}
rewards a card whose best detector scores well, whose detectors were broadly evaluated ($\mathcal D_s\subseteq\mathcal D$ evaluated, $v^{\max}_s=\max_m v_{s,m}$), and whose ranking is decisive. The top eight cards form the neighbor set $\mathcal{N}_X$.

\noindent\textbf{From cards to a calibrated prior.} Raw VUS-PR is not comparable across time series, so each card's results are rank-normalized: the best detector on $s$ receives $u_{s,m}=1$, the worst $0$, and the others are spaced evenly. Each selected card is weighted by $\beta_s=\mathrm{sim}_s\,(0.5+0.5\,\lambda_s)$, and $\hat\beta_s=\beta_s/\sum_{s'\in\mathcal{N}_X}\beta_{s'}$ denotes the normalized weight. For every detector, the prior reports a weighted utility and its dispersion,
\begin{align*}
U_m&=\sum_{s\in\mathcal{N}_X}\hat\beta_s\,u_{s,m},\\
\sigma_m&=\Big(\sum_{s\in\mathcal{N}_X}\hat\beta_s\,(u_{s,m}-U_m)^2\Big)^{1/2},
\end{align*}
together with its support (the number of cards that evaluated it) and effective support $\exp(H(\hat{\boldsymbol\beta}))$, where $H$ is the entropy of the normalized weight vector $\hat{\boldsymbol\beta}$. The prior also lists, for each detector, the cards on which it ranks in the top three (supporting) or last (contradictory).

The prior carries a retrieval confidence $\gamma_X\in[0,1]$:
\begin{align*}
\gamma_X ={}& 0.30\,\mathrm{agree} + 0.25\,\mathrm{mass}\\
&+ 0.20\,\mathrm{even} + 0.25\,(1-\mathrm{ood}).
\end{align*}
Here \emph{agree} is the mean pairwise Jaccard overlap of the selected cards' top-three detectors, \emph{mass} is the combined weight share of the three highest-utility detectors (clipped to one), \emph{even} is Pielou's evenness of the source datasets among the selected cards, and \emph{ood} is the normalized distance to the nearest card (clipped to one). Confidence is low when neighbors disagree, when one source dominates, or when the query lies far from every card.

\noindent\textbf{How the prior is used.} The planner receives the prior in its prompt: the ranked utilities, their uncertainties and support, and $\gamma_X$ with its components, so it can discount a ranking that rests on weak or conflicting precedents. The prior strength $\kappa_X=\kappa_0\gamma_X$ (with $\kappa_0=1$) also sets the benchmark confidence of every contract entry, one of the quantities read by the semantic-scorer gate (Table~\ref{tab:fixed-method-config}). The prior shapes only which candidates are proposed: it is never used as a label, and ensemble weights are computed from current-run validation alone (Section~\ref{sec:sphinx-validation}).

\noindent\textbf{Example.} For the SMAP time series of Table~\ref{tab:contract-examples}, the eight retrieved cards give KMeansAD the highest utility ($U_m=0.90$, $\sigma_m=0.24$), followed by KNN ($0.68$, $0.14$) and DeepAnt ($0.58$), with $\gamma_X=0.52$. Neighbors agree on the top detector but not on the rest of the ranking (agreement $0.24$). The final ensemble for this time series contains KMeansAD, KNN, PCA, and IForest; KNN and PCA are among the prior's top four, while IForest was admitted on the validation evidence alone.

\subsection{Semantic Scorers}
\label{app:semantic-scorers}

\noindent\textbf{What a semantic scorer is.} Detector cards describe fixed, offline detectors whose inductive biases are set by their designers. A semantic scorer is instead built by the LLM during the run, for one entry of the frozen contract. It turns the entry's operational description and expected score behavior into a small program of deterministic score operators. The planner sees a catalog of \emph{semantic biases} (Table~\ref{tab:semantic-operators}), not a second detector menu. Each bias names a behavior, such as ``deviation from a local contextual baseline'' or ``sharp local curvature'', and is implemented by exactly one label-free operator.

\begin{table*}[t]
\centering
\caption{Examples of semantic biases available to the planner when it constructs a semantic scorer. Each bias is implemented by one deterministic, label-free score operator; the planner may set only the listed parameters.}
\label{tab:semantic-operators}
\small
\begin{tabular}{@{}llll@{}}
\toprule
\textbf{Semantic bias} & \textbf{Score operator} & \textbf{Planner-set parameters} & \textbf{Targeted behavior} \\
\midrule
Robust tail & Robust $z$-score & -- & Point / tail outliers \\
Local context residual & Rolling-median residual & window & Local contextual deviations \\
Seasonal phase residual & Same-phase peer residual & period & Seasonal phase breaks \\
Abrupt change & First difference & -- & Sudden point-to-point change \\
Curvature impulse & Second difference & -- & Sharp impulses, point shocks \\
Multiscale context & Multiscale residual & windows & Multi-scale local deviations \\
Adaptive baseline & EWMA residual & span & Adaptive baseline deviations \\
Regime shift & Left/right changepoint score & window & Mean / regime changes \\
Volatility shift & Rolling-variance shift & window, direction & Noisy bursts, volatility changes \\
Plateau stuckness & Plateau score & window, tolerances & Stuck sensors, flat plateaus \\
\bottomrule
\end{tabular}
\end{table*}

\noindent\textbf{Specification.} A scorer proposal declares the targeted contract entry, the rationale, and the expected signal. It lists one to four biases, each with an operator, the operator's parameters, and a weight. It also gives a preprocessing plan and a sequence of score transforms drawn from a fixed set: rank normalization, rank power, min--max scaling, and an event envelope that spreads or sustains a peak over a short window. The parser drops unknown operators, transforms, and parameters, and a malformed specification falls back to a robust $z$-score, so a proposal can fail only softly and always remains auditable.

\noindent\textbf{Execution.} Each operator produces one score stream. Streams are min--max scaled and combined by their normalized weights, and the declared transforms are then applied in order. Every intermediate stream is stored with the candidate, so the final score can be traced back to the operators and parameters that produced it.

\noindent\textbf{Gate.} Scorers are a complement, not a replacement, for the offline detectors. The gate therefore requires that at least one detector candidate has already run in the episode, and that the targeted semantic can be compiled into the operator catalog. It also requires remaining budget and one of four triggers: the semantic's coverage is below $0.5$, no validated detector supports it yet, its benchmark confidence is below $0.5$, or two or more redundant candidates already target it (Table~\ref{tab:fixed-method-config}). The trigger that fired is recorded in the evidence bank. An admitted scorer is then validated on the same shared syllabus as every detector and enters the ensemble only through the same evidence-calibrated aggregation.

\noindent\textbf{Usage in the reported runs.} Detectors from the catalog covered most contracts. Across the univariate evaluation time series, the planner built a semantic scorer in $5$ of $113$ episodes with a stored trace, always through the low-coverage trigger, and each built scorer was retained in the final ensemble. It built none on the multivariate time series. Appendix~\ref{app:scorer-cases} shows two time series on which a scorer improves the final ensemble. Two of the scorers built in the reported runs follow.

\noindent\textbf{Example 1: a point-spike scorer (UCR \#79, Human Activity).} After three detectors had covered the contextual, collective, and level-shift entries, the point-spike entry (AS4) remained uncovered. The planner reasoned that the time series is very smooth and highly autocorrelated (lag-one autocorrelation $0.99$, noise $0.14$), so a narrow impulse is distinct from the patterns already tested, and that a local-residual and difference scorer fits this morphology better than another catalog detector. It proposed
\begin{center}
\small
$0.35\cdot\text{rolling-median residual}(w{=}11) + 0.30\cdot\text{second difference} + 0.20\cdot\text{first difference} + 0.15\cdot\text{robust } z$,
\end{center}
followed by rank normalization, rank power $1.4$, and min--max scaling. In validation it localized injected spikes well (localization $0.91$, normal-region score $0.84$, score shape $0.74$) and raised AS4 coverage from $0$ to $0.74$. The final ensemble is LSTMAD, KMeansAD, this scorer, and DeepAnt.

\noindent\textbf{Example 2: a level-shift scorer (SVDB \#1, Medical).} Three detectors had given weak or mixed support for the first three entries, and the level-shift entry was uncovered. The planner cited strong seasonality ($0.65$) with almost no global trend ($0.002$): a persistent displacement of the baseline is then distinguishable from the normal period-8 oscillation. It proposed
\begin{center}
\small
$0.45\cdot\text{changepoint}(w{=}8) + 0.35\cdot\text{EWMA residual}(\text{span}{=}16) + 0.20\cdot\text{same-phase residual}(\text{period}{=}8)$,
\end{center}
followed by robust rank normalization and a sustaining event envelope of width $8$, so that the score stays high over a shifted segment instead of peaking only at its boundary. Validation confirmed the intended shape (score shape $0.97$, localization $1.00$) but found weak recovery of injected shifts ($0.26$), and the scorer was recorded with a weak-proxy warning. It still raised level-shift coverage to $0.72$ and joined the final ensemble with LSTMAD, HBOS, and KMeansAD, with a weight set by its validation evidence.

Both scorers were composed from the same catalog as generic building blocks. The LLM contributes the choice of operators, their weights, and their windows, tied to the profiled period and smoothness of the time series, and every choice is recorded as evidence that can be inspected.

\subsection{Experimental Setup Details}
\label{app:setup-details}

\noindent\textbf{Detector catalog.} Every detector is available in both benchmark settings, so the univariate and multivariate settings share the same model inventory. For a candidate-model action, \system{} may change only registered hyperparameters, and unknown parameters are discarded before execution. The coding agents receive the catalog in \texttt{DETECTORS.md} and may select, configure, or ensemble its detectors in their generated solution.

\noindent\textbf{Baselines.} Each baseline follows its original publication; LLMAD, MLLMAD, Codex, and AD-AGENT use GPT-5.5.
\begin{itemize}[noitemsep,topsep=2pt,parsep=0pt,partopsep=0pt,leftmargin=0.4cm]
    \item \emph{Global Best}~\citep{liu2025tsb} applies to every time series the single detector with the highest mean VUS-PR over the evaluation set. \emph{Supervised Selection}~\citep{liu2025tsb} uses each evaluation time series' domain label and applies the detector that performs best on the training time series of that domain. Both consult labels or curated metadata and serve only as references.
    \item \emph{Random Ensemble} averages the outputs of five candidates drawn at random.
    \item \emph{AvgEns}~\citep{aggarwal2015theoretical} standardizes the scores of all candidates and averages them per timestamp, so it commits to no single detector.
    \item \emph{MetaOD}~\citep{zhao2021automatic} treats model selection as collaborative filtering. It factorizes the performance matrix of training time series by detectors, learns a map from a time series' meta-features to its latent factors, and selects the detector with the highest predicted performance on the new time series.
    \item \emph{MSAD}~\citep{sylligardos2023choose} casts selection as time-series classification. A classifier trained on windows of the training time series predicts the best detector for each window, and a majority vote over the windows of a new time series picks its detector.
    \item \emph{TSADAMS}~\citep{goswami2023unsupervised} selects a detector without labels. It ranks candidates by several surrogate metrics, including forecasting error, model centrality, and performance on injected synthetic anomalies, and merges the rankings by Borda rank aggregation.
    \item \emph{AutoOD}~\citep{cao2023autood} generates a new detector rather than selecting one. It derives pseudo-labels from the agreement of the candidate detectors, here by majority vote, and iteratively trains a classifier on the reliably labeled points to score the time series.
    \item \emph{LLMAD} and \emph{MLLMAD}~\citep{park2025delving} prompt an LLM zero-shot to return anomalous intervals. LLMAD gives it each de-seasonalized window as indexed text, and MLLMAD adds a line plot of the same window.
    \item \emph{MLLM4TS}~\citep{liu2026mllmts} renders a multivariate time series as an image and fuses vision-encoder features with time-series patch embeddings in a language-model backbone. Trained to reconstruct normal behavior, it scores each timestamp by its reconstruction error.
    \item \emph{Claude Code}~\citep{anthropic_claude_2026} and \emph{Codex}~\citep{openai_codex_2026} are general-purpose coding agents. Given the time series and the detector catalog, each writes and runs its own detection program.
    \item \emph{AD-AGENT}~\citep{yang2025ad} is a multi-agent anomaly-detection framework. Its agents parse the request, select a detector from existing libraries, gather documentation, generate and debug code, and evaluate the result.
\end{itemize}

All experiments, including baseline comparisons, use a fixed random seed for reproducibility. Evaluation across 146 time-series datasets provides a broad empirical basis for assessing performance.

\subsection{Dataset Descriptions}
\label{app:dataset-descriptions}

\noindent\textbf{Dataset restriction.} The TSB-AD collections~\citep{liu2024elephant} were introduced to address label noise, dataset bias, and feasibility constraints in earlier TSAD benchmarks~\citep{wu2021current,liu2024elephant}. Licensing terms prevent us from using every source dataset, so we keep the twelve we are permitted to use. The restriction is a licensing constraint alone. It is unrelated to time series difficulty, length, dimensionality, or detector behavior, and every retained time series is used exactly as TSB-AD releases it. The retained benchmark has 377 univariate and 114 multivariate time series. The sources span medical, facility, human-activity, sensor, synthetic, and environmental data, and they cover real and injected point, contextual, and collective anomalies. The comparison therefore remains a heterogeneous multi-domain test. We preserve the domain-aware assignment of TSB-AutoAD~\citep{liu2025tsb} for every retained time series, so the meta-training and evaluation sets remain disjoint. The evaluation set is used without its labels by every deployable method; labels enter only the metric computation and the label-consulting Baseline Solution references.

% Auto-generated by draft/exp_results/build_dataset_table.py.
\begin{table*}[t]
\centering
\caption{Domain-wise statistics and splits for the twelve license-permitted TSB-AD source datasets.}
\label{tab:benchmark_datasets}
{
\begin{tabular}{lccccl}
\toprule
\textbf{Domain}
& \multicolumn{2}{c}{\textbf{Univariate}}
& \multicolumn{2}{c}{\textbf{Multivariate}}
& \multirow{2}{*}{\textbf{Source datasets}} \\
\cmidrule(lr){2-3} \cmidrule(lr){4-5}
& \textbf{Total} & \textbf{Train / Eval}
& \textbf{Total} & \textbf{Train / Eval} & \\
\midrule
Medical         & 144       & 104 / 40  & 49        & 36 / 13   & LTDB, MITDB, SVDB, UCR \\
Facility        & 101       & 70 / 31   & 22        & 16 / 6    & SMD, UCR \\
Human Activity  & 58        & 41 / 17   & 9         & 6 / 3     & Daphnet, OPP, UCR \\
Sensor          & 34        & 25 / 9    & 34        & 25 / 9    & CATSv2, GECCO, SMAP, UCR \\
Synthetic       & 24        & 12 / 12   & --        & --        & MGAB, TODS \\
Environment     & 16        & 10 / 6    & --        & --        & UCR \\
\midrule
\textbf{Total}  & 377       & 262 / 115 & 114       & 83 / 31   & \\
\bottomrule
\end{tabular}
}
\end{table*}

\noindent\textbf{Source datasets.} The twelve retained source datasets are described below.

\begin{itemize}[noitemsep,topsep=0pt,parsep=0pt,partopsep=0pt,leftmargin=0.5cm]
    \item \textbf{UCR~\citep{wu2021current}} is a collection of univariate time series from multiple domains, including air temperature, arterial blood pressure (ABP), astronomy, EPG, ECG, gait, power demand, respiration, and walking acceleration. Most anomalies are introduced artificially.
    \item \textbf{MGAB~\citep{markus_thill_2020_3762385}} comprises Mackey--Glass time series whose anomalies exhibit chaotic behavior that is difficult to distinguish visually.
    \item \textbf{TODS~\citep{lai2021revisiting}} is a synthetic dataset comprising global, contextual, shapelet, seasonal, and trend anomalies.
    \item \textbf{Daphnet~\citep{bachlin2009wearable}} contains annotated readings from three acceleration sensors placed on the hip and leg of patients with Parkinson's disease who experience freezing of gait during walking tasks.
    \item \textbf{OPPORTUNITY~\citep{roggen2010collecting}} (OPP) was devised to benchmark human-activity recognition algorithms, including classification, automatic data segmentation, sensor fusion, and feature extraction. It comprises motion-sensor readings recorded while users performed typical daily activities.
    \item \textbf{SMD~\citep{10.1145/3292500.3330672}} is a five-week dataset collected from a large Internet company and contains three groups of entities from 28 different machines.
    \item \textbf{SMAP~\citep{hundman2018detecting}} contains real spacecraft telemetry from the Soil Moisture Active Passive satellite. Each time series has one feature representing a sensor measurement, while the remaining features represent binary-encoded commands.
    \item \textbf{GECCO~\citep{moritz2018gecco}} is a water-quality dataset used in a competition on online anomaly detection for drinking-water quality.
    \item \textbf{MITDB~\citep{goldberger2000physiobank}} contains 48 half-hour excerpts of two-channel ambulatory ECG recordings obtained from 47 subjects studied by the BIH Arrhythmia Laboratory between 1975 and 1979.
    \item \textbf{SVDB~\citep{greenwald_improved_1990}} includes 78 half-hour ECG recordings selected to supplement examples of supraventricular arrhythmias in the MIT-BIH Arrhythmia Database.
    \item \textbf{LTDB~\citep{goldberger2000physiobank}} is a collection of seven long-duration ECG recordings, each lasting 14--22 hours, with manually reviewed beat annotations.
    \item \textbf{CATSv2~\citep{fleith2023cats}} is the second version of the Controlled Anomalies Time Series dataset. It consists of commands, external stimuli, and telemetry readings from a simulated complex dynamical system with 200 injected anomalies.
\end{itemize}

\begin{table*}[t]
\centering
\caption{The 15-model candidate set shared by \system{} and the coding-agent baselines. All models are exposed in both the univariate (U) and multivariate (M) experiments.}
\label{tab:candidate}
\small
{
\begin{tabular}{llccl}
\toprule
\textbf{Family} & \textbf{Candidate Model} & \textbf{U} & \textbf{M} & \textbf{Score principle} \\
\midrule
\multirow{8}{*}{Statistical}
& MCD~\cite{rousseeuw1999fast}                    & \ding{51} & \ding{51} & Robust covariance deviation \\
& OCSVM~\cite{scholkopf_support_1999}             & \ding{51} & \ding{51} & One-class decision boundary \\
& LOF~\cite{breunig2000lof}                       & \ding{51} & \ding{51} & Local-density deviation \\
& KNN~\cite{ramaswamy2000efficient}               & \ding{51} & \ding{51} & Nearest-neighbor distance \\
& KMeansAD~\cite{yairi2001fault}                  & \ding{51} & \ding{51} & Distance to cluster centroid \\
& IForest~\cite{liu2008isolation}                 & \ding{51} & \ding{51} & Isolation depth \\
& HBOS~\cite{goldstein2012histogram}              & \ding{51} & \ding{51} & Histogram-density score \\
& PCA~\cite{aggarwal2017introduction}             & \ding{51} & \ding{51} & Principal-subspace deviation \\
\midrule
\multirow{7}{*}{Neural}
& AutoEncoder~\cite{sakurada2014anomaly}          & \ding{51} & \ding{51} & Reconstruction error \\
& LSTMAD~\cite{LSTManomaly}                       & \ding{51} & \ding{51} & LSTM forecasting error \\
& DeepAnt~\cite{munir2018deepant}                 & \ding{51} & \ding{51} & Convolutional forecasting error \\
& Donut~\cite{xu2018unsupervised}                 & \ding{51} & \ding{51} & Variational reconstruction score \\
& OmniAnomaly~\cite{10.1145/3292500.3330672}      & \ding{51} & \ding{51} & Stochastic-RNN reconstruction score \\
& USAD~\cite{audibert2020usad}                    & \ding{51} & \ding{51} & Adversarial reconstruction error \\
& TranAD~\cite{tuli2022tranad}                    & \ding{51} & \ding{51} & Transformer reconstruction error \\
\bottomrule
\end{tabular}
}
\end{table*}

\section{Additional Experimental Results}
\label{app:extra-results}

This appendix reports per-dataset and per-domain results, token accounting, and runtime, and reproduces artifacts from the reported runs: semantic contracts, an end-to-end run trajectory, a complete detection report, and domain-context hypotheses.

\subsection{Per-Dataset Results}
\label{app:per-dataset}

Table~\ref{tab:per-dataset} reports the source-level VUS-PR underlying Table~\ref{tab:main-comparison}, highlighting \system{}'s broad coverage. On univariate time series, \system{} leads on UCR, SMD, and MGAB, which together account for $83$ of $115$ evaluation series. Its advantage on MGAB is particularly pronounced: $0.5977$, compared with at most $0.2613$ for any baseline. On multivariate time series, \system{} leads on SMD, LTDB, and CATSv2. It therefore achieves the highest score on more source datasets than any baseline in either benchmark.

The remaining dataset leaders vary: MSAD leads on univariate OPP and LTDB, AvgEns on SVDB and SMAP, and Codex on TODS and MITDB; for multivariate series, MSAD leads on SMAP, Claude Code on SVDB and MITDB, and AD-AGENT on OPP. No baseline leads on more than two datasets in either benchmark. Since several sources contain only one to three series, these estimates are complemented by the domain-level analysis in Appendix~\ref{app:domain-consistency}.

% Auto-generated by draft/exp_results/build_per_dataset_table.py.
% Regenerate: python draft/exp_results/build_per_dataset_table.py
\begin{table*}[t]
\centering
\caption{VUS-PR per source dataset on the license-permitted evaluation time series, supplementing Table~\ref{tab:main-comparison}. $n$ is the number of evaluation time series from the dataset; rows are ordered by $n$, and \emph{All} repeats the per-series means of Table~\ref{tab:main-comparison}. \textbf{Bold} marks the best method in each row; the label-consulting Baseline Solution entries are omitted. -- marks a method not run on that benchmark. Datasets with one or two series are noisy and should be read with care.}
\label{tab:per-dataset}
\footnotesize
\setlength{\tabcolsep}{3pt}
\resizebox{\textwidth}{!}{%
\begin{tabular}{@{}lrcccccccccccc@{}}
\toprule
\textbf{Dataset} & $n$ & \textbf{AvgEns} & \textbf{MetaOD} & \textbf{MSAD} & \textbf{TSADAMS} & \textbf{AutoOD} & \textbf{LLMAD} & \textbf{MLLMAD} & \textbf{MLLM4TS} & \textbf{Claude Code} & \textbf{Codex} & \textbf{AD-AGENT} & \textbf{\system{}} \\
\midrule
\multicolumn{14}{@{}l}{\emph{Univariate}} \\
\midrule
UCR & 64 & 0.3561 & 0.2849 & 0.3230 & 0.2200 & 0.0196 & 0.0636 & 0.0491 & -- & 0.2455 & 0.2274 & 0.3244 & \textbf{0.3910} \\
SMD & 15 & 0.5683 & 0.2362 & 0.1412 & 0.1771 & 0.3962 & 0.1485 & 0.1714 & -- & 0.5188 & 0.5303 & 0.1426 & \textbf{0.7710} \\
SVDB & 8 & \textbf{0.6052} & 0.4019 & 0.3175 & 0.0728 & 0.1088 & 0.1744 & 0.2054 & -- & 0.4224 & 0.3274 & 0.3009 & 0.3532 \\
TODS & 8 & 0.6873 & 0.6536 & 0.6088 & 0.6779 & 0.5351 & 0.7800 & 0.7693 & -- & 0.7882 & \textbf{0.8661} & 0.7748 & 0.8187 \\
OPP & 7 & 0.6914 & 0.4925 & \textbf{0.8285} & 0.2200 & 0.7236 & 0.1332 & 0.1521 & -- & 0.3613 & 0.1875 & 0.2553 & 0.7221 \\
SMAP & 5 & \textbf{0.8677} & 0.6599 & 0.3897 & 0.2144 & 0.1543 & 0.1241 & 0.1082 & -- & 0.6996 & 0.5465 & 0.5295 & 0.8549 \\
MGAB & 4 & 0.1310 & 0.2613 & 0.2613 & 0.0059 & 0.0057 & 0.0031 & 0.0061 & -- & 0.0120 & 0.0153 & 0.1598 & \textbf{0.5977} \\
LTDB & 3 & 0.7479 & 0.6474 & \textbf{0.8230} & 0.2604 & 0.3777 & 0.2910 & 0.3482 & -- & 0.5406 & 0.6112 & 0.4560 & 0.6905 \\
MITDB & 1 & 0.1460 & 0.0449 & 0.0425 & 0.0456 & 0.0506 & 0.0465 & 0.0469 & -- & 0.2374 & \textbf{0.4475} & 0.0610 & 0.1777 \\
\midrule
\emph{All} & 115 & 0.4673 & 0.3479 & 0.3609 & 0.2278 & 0.1687 & 0.1428 & 0.1412 & -- & 0.3575 & 0.3343 & 0.3305 & \textbf{0.5212} \\
\midrule
\multicolumn{14}{@{}l}{\emph{Multivariate}} \\
\midrule
SVDB & 10 & 0.1761 & 0.2198 & 0.2358 & 0.0610 & 0.0769 & -- & -- & 0.0810 & \textbf{0.2900} & 0.1410 & 0.0877 & 0.2489 \\
SMAP & 8 & 0.2081 & 0.3186 & \textbf{0.3806} & 0.1539 & 0.1518 & -- & -- & 0.2844 & 0.2337 & 0.1548 & 0.1347 & 0.2372 \\
SMD & 6 & 0.6174 & 0.3511 & 0.3212 & 0.2114 & 0.2916 & -- & -- & 0.5809 & 0.1810 & 0.3339 & 0.5155 & \textbf{0.6176} \\
OPP & 3 & 0.2677 & 0.0967 & 0.3827 & 0.0930 & 0.2540 & -- & -- & 0.1486 & 0.2119 & 0.2810 & \textbf{0.3887} & 0.2799 \\
LTDB & 2 & 0.3800 & 0.3412 & 0.4500 & 0.3093 & 0.2936 & -- & -- & 0.3415 & 0.3185 & 0.0981 & 0.3100 & \textbf{0.5765} \\
CATSv2 & 1 & 0.2382 & 0.2508 & 0.2508 & 0.0305 & 0.1104 & -- & -- & 0.1465 & 0.0000 & 0.1031 & 0.2909 & \textbf{0.4131} \\
MITDB & 1 & 0.0771 & 0.0828 & 0.0828 & 0.0510 & 0.0543 & -- & -- & 0.0731 & \textbf{0.1802} & 0.0581 & 0.0000 & 0.1414 \\
\midrule
\emph{All} & 31 & 0.2906 & 0.2632 & 0.3133 & 0.1319 & 0.1693 & -- & -- & 0.2554 & 0.2358 & 0.1888 & 0.2298 & \textbf{0.3432} \\
\bottomrule
\end{tabular}%
}
\end{table*}

\subsection{Consistency Across Domains}
\label{app:domain-consistency}

Since licensing restrictions exclude entire source datasets, we assess the breadth of \system{}'s aggregate gains across the retained domains. We group evaluation time series according to Table~\ref{tab:benchmark_datasets} and compare against all baselines.

On univariate time series, \system{} achieves the highest VUS-PR in three of six domains, covering $83$ of $115$ series: Medical ($0.5084$ versus $0.4587$ for AvgEns), Facility ($0.4761$ versus $0.3632$), and Synthetic ($0.7450$ versus $0.5825$ for Codex). Other baselines show complementary strengths in smaller domains: AvgEns leads on Environment ($n=6$; $0.7446$ versus $0.4016$ for \system{}) and Sensor ($n=9$; $0.8266$ versus $0.7365$), while MSAD leads on Human Activity ($n=17$; $0.4223$ versus $0.4034$).

On multivariate time series, \system{} leads on Facility and Medical, covering $19$ of $31$ series, with margins of $0.0002$ over AvgEns and $0.0050$ over Claude Code, respectively. MSAD leads on Sensor ($n=9$; $0.3662$ versus $0.2568$ for \system{}), and AD-AGENT on Human Activity ($n=3$; $0.3887$ versus $0.2799$). Across these domains, \system{} maintains a balanced performance profile: its largest gap to a domain's best method is $0.109$ VUS-PR, smaller than those of AvgEns ($0.155$), AD-AGENT ($0.214$), and MSAD ($0.296$), whose Facility score is $0.3212$.

Together, these results support \system{}'s broad effectiveness. Ensemble pooling and historical selection offer complementary strengths in individual domains, while \system{} combines broad coverage with a smaller worst-domain gap.

\subsection{Token Accounting}
\label{app:token-accounting}

The token counts in Figure~\ref{fig:token-cost} are the total language-model tokens (prompt plus completion) consumed per evaluation time series, averaged over each split. For \system{}, AD-AGENT, and the coding agents, this covers every call made during the episode; for LLMAD and MLLMAD, it covers the calls that submit the windows of the time series to the model.

\subsection{Runtime}
\label{app:runtime}

Table~\ref{tab:runtime} reports the mean wall-clock time per univariate evaluation time series. For agent-based methods, it is the time of the whole episode, from receiving the time series to returning the final scores. For AvgEns, it is the total runtime of all candidate detectors it executes. For LLMAD and MLLMAD, it is the per-call latency multiplied by the number of windows per time series.

\begin{table}[t]
\centering
\caption{Mean wall-clock time per univariate evaluation time series ($n=115$). Appendix~\ref{app:runtime} describes how each time is measured.}
\label{tab:runtime}
\small
\begin{tabular}{@{}l r@{}}
\toprule
\textbf{Method} & \textbf{Wall-clock (s)} \\
\midrule
AvgEns & 747.3 \\
LLMAD & 245 \\
MLLMAD & 314 \\
Claude Code & 130.3 \\
Codex & 378.7 \\
AD-AGENT & 251.0 \\
\midrule
\system{} & 301.1 \\
\bottomrule
\end{tabular}
\end{table}

\subsection{Case Study: Semantic Scorers}
\label{app:scorer-cases}

\begin{figure*}[t]
\centering
\includegraphics[width=\textwidth]{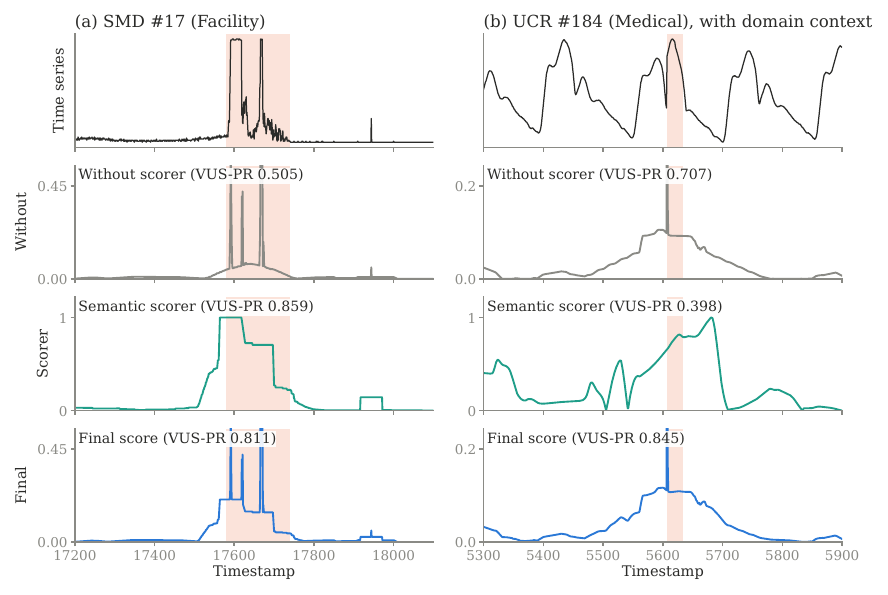}
\vspace{-0.5cm}
\caption{How a semantic scorer changes the final anomaly score. Rows, top to bottom: the time series, with labeled anomalies shaded; the ensemble score without the scorer; the scorer's own score; and the final score of the ensemble that includes it. Each score is min--max normalized over the whole series, and its VUS-PR is computed on the whole series. The views are zoomed to the anomaly. In the second and fourth rows, the score axis is clipped so that the sustained response is visible beneath the highest peaks. (a)~The planner built the scorer during the run without benchmark cards. (b)~The scorer was proposed from the domain context and admitted by the benefit gate.}
\label{fig:scorer-cases}
\end{figure*}

Figure~\ref{fig:scorer-cases} shows two time series to study the impact of the semantic scorer (Appendix~\ref{app:semantic-scorers}). For each, we show the score of the ensemble without the scorer, the scorer's own score, and the final score that combines them.

\noindent\textbf{(a) A planner-built level-shift scorer (SMD \#17, Facility).} The time series is strongly seasonal (seasonality $0.92$, lag-one autocorrelation $0.95$, cycle length about $55$). Its labeled anomaly is a segment of about $160$ points in which the level rises and stays high, with several spikes. In the run without benchmark cards, the planner first proposed DeepAnt, KMeansAD, and HBOS for the contextual, collective, and point-spike entries. Validation flagged DeepAnt as unstable under perturbation, KMeansAD as weak on synthetic recovery, and HBOS as failing to localize spikes, and the level-shift entry remained uncovered. Through the low-coverage trigger, the planner composed
\begin{center}
\small
$0.45\cdot\text{same-phase residual}(\text{period}{=}55) + 0.40\cdot\text{changepoint}(w{=}55) + 0.15\cdot\text{EWMA residual}(\text{span}{=}55)$,
\end{center}
followed by a sustaining event envelope of width $55$. It argued that a persistent offset should be measured against a same-phase seasonal baseline rather than a global threshold. On its own, the scorer is the strongest candidate in the ensemble ($0.859$ VUS-PR): it rises at the onset of the shift and stays high across the shifted segment (third row of Figure~\ref{fig:scorer-cases}a). It covered the level-shift entry at $0.55$ and received $26\%$ of the ensemble weight. Without it, the final score peaks at the spikes but falls back between them: $41\%$ of the anomalous timestamps score below the $99$th percentile of normal timestamps. The scorer keeps the score raised across the segment, reducing this share to $26\%$, and VUS-PR rises from $0.505$ to $0.811$.

\noindent\textbf{(b) A domain-context variance scorer (UCR \#184, Medical).} The time series is a near-periodic signal (seasonality $0.99$, cycle length $153$), and its labeled anomaly is a single distorted cycle of about $27$ points. The no-context ensemble of KMeansAD, KNN, LSTMAD, and HBOS already reaches $0.707$ VUS-PR. It responds to the distorted cycle with a narrow peak at its onset, but its response over the rest of the cycle is barely higher than that of neighboring normal cycles. With the domain description, the extra LLM call proposed two scorers. The first, a cycle-template residual for the contextual entry, was held out by the benefit gate because of weak synthetic recovery (coverage $0.31$). The second targets the entry for a temporary loss of periodic regularity with a rolling-variance shift over one cycle ($w{=}153$). It covered that entry at $0.78$, raised the proxy score from $0.746$ to $0.886$, and was admitted with weight $0.25$. On its own, the scorer is a weak detector ($0.398$ VUS-PR): its response is broad, and it peaks shortly after the distorted cycle because the variance window spans a full cycle (third row of Figure~\ref{fig:scorer-cases}b). Its value is complementary. Combined with the base ensemble, it lifts the final score across the whole distorted cycle, so fewer normal timestamps outrank the anomaly: the share that exceeds the anomaly's median score falls from $0.45\%$ to $0.25\%$, and VUS-PR rises to $0.845$.

\noindent\textbf{Takeaways.} In both cases, the catalog detectors respond to the sharpest part of the anomaly but not to its full extent, and the scorer supplies the missing segment-level response that its contract entry describes. A scorer need not be a strong detector on its own to help: in (b), a weak scorer still improves the ensemble because it responds where the detectors do not. Conversely, in (a) the final score ($0.811$) remains slightly below the scorer alone ($0.859$): the label-free weights cannot tell which candidate is best on this series, so the ensemble trades a small loss against its strongest member for robustness to any single candidate. Because each scorer forms its own semantic group, the cross-group maximum (Eq.~\eqref{eq:semantic-ensemble}) adds this response without diluting the detectors' peaks. Each scorer is a short, inspectable program whose operators and windows are tied to the profiled cycle length, so the source of the improvement can be read directly from the evidence bank.

\subsection{Semantic Contract Examples}
\label{app:contract-examples}

Table~\ref{tab:contract-examples} reproduces the frozen semantic contracts that the grounding agent produced for two evaluation time series in the reported runs, without domain context. Entries are listed by priority; priorities within a contract sum to one. Both contracts use the same five-type vocabulary, but the profiles lead to different targets and orderings.

\noindent\textbf{Telemetry (SMAP \#17, multivariate).} The time series has $7{,}331$ points and $25$ channels, with weak serial dependence (lag-one autocorrelation $0.06$), weak seasonality ($0.15$), high noise ($0.73$), a heavy right tail (skewness $4.1$, kurtosis $31.1$), and low mean absolute cross-channel correlation ($0.07$). The contract therefore puts point spikes first and collective bursts second, and ranks correlation change last because the channels are nearly independent.

\noindent\textbf{ECG (MITDB \#8, univariate).} The time series has $650{,}000$ points in one channel, with strong seasonality ($0.85$), strong serial dependence ($0.96$), and a dominant period near $25$ samples. Here the contract puts contextual anomalies first, meaning values at the wrong phase of the cycle, and collective disruptions of the cycle second. Point spikes drop to third, behind the two entries that describe departures from the cycle. No correlation-change entry appears for a single channel; a second collective entry targets changes in noise texture instead.

Each entry carries three validation questions, which the syllabus designer turns into synthetic interventions and score checks (Section~\ref{sec:sphinx-validation}). For the top two entries of each contract they read:

\begin{itemize}[noitemsep,topsep=2pt,parsep=0pt,partopsep=0pt,leftmargin=0.4cm]
    \item \textbf{SMAP AS1 (point spike).} \emph{Injection recovery:} ``Does the score peak when isolated high-amplitude positive spikes are injected relative to each channel's local noise scale?'' \emph{Localization:} ``Are score maxima tightly localized to the spike samples rather than spread across long surrounding regions?'' \emph{Normal region:} ``Do high scores remain rare in typical noisy regions without extreme values?''
    \item \textbf{SMAP AS2 (collective).} \emph{Injection recovery:} ``Does the score rise over injected bursts of repeated moderate outliers or increased variance lasting multiple samples?'' \emph{Score shape:} ``Is the high-score region sustained across most of the abnormal run and lower outside it?'' \emph{Normal region:} ``Can the detector avoid flagging ordinary noisy windows that preserve the baseline distribution and short-run structure?''
    \item \textbf{MITDB AS1 (contextual).} \emph{Injection recovery:} ``If a seasonal cycle is locally phase-shifted, amplitude-distorted, or has values injected at the wrong phase, does the score rise within that affected cycle?'' \emph{Localization:} ``Does the score concentrate around the phase-inconsistent samples rather than flagging the entire seasonal waveform?'' \emph{Normal region:} ``On clean high-autocorrelation seasonal regions with normal amplitude variation, does the detector avoid repeated false positives?''
    \item \textbf{MITDB AS2 (collective).} \emph{Injection recovery:} ``If one or more windows have the seasonal pattern flattened, skipped, or made irregular for about one cycle or longer, does the detector recover the anomalous segment?'' \emph{Score shape:} ``Are scores elevated across the full structural disruption rather than only at its first extreme point?'' \emph{Localization:} ``Does the detected interval align with the disrupted cycles to within roughly one seasonal period?''
\end{itemize}

\begin{table*}[t]
\centering
\caption{Two frozen semantic contracts from the reported runs (no domain context), reproduced verbatim. Prio.\ is the entry priority; shape is the expected event shape.}
\label{tab:contract-examples}
\footnotesize
\setlength{\tabcolsep}{3pt}
\renewcommand{\arraystretch}{1.1}
\resizebox{\textwidth}{!}{%
\begin{tabular}{@{}l l c l p{0.47\textwidth} p{0.24\textwidth}@{}}
\toprule
\textbf{ID} & \textbf{Type} & \textbf{Prio.} & \textbf{Shape} & \textbf{Operational description} & \textbf{Expected score behavior} \\
\midrule
\multicolumn{6}{@{}l}{\emph{SMAP \#17 (multivariate): 25 channels, weak autocorrelation, heavy right tail, low cross-channel correlation}} \\
AS1 & point spike & 0.34 & point & Isolated extreme observations in one or more channels, especially large positive excursions, are anomalous because the value distribution is strongly right-skewed and heavy-tailed while serial dependence is weak. & Sharp peaks at isolated extreme samples with rapid return to baseline. \\
AS2 & collective & 0.28 & segment & Short runs of unusual volatility, repeated outliers, or abnormal motif/transition structure are anomalous even when individual samples are not uniquely extreme; local features show long stretches and burst-like structure despite weak global autocorrelation. & Sustained high over the abnormal run, with elevated segment-level score rather than a single-sample peak. \\
AS3 & contextual & 0.18 & contextual & Values or local patterns that are plausible marginally but abnormal for their local channel context are anomalous, reflecting weak seasonality, moderate nonstationarity, and high noise where context-sensitive deviations matter more than global thresholds alone. & Localized elevation where the observation or short pattern departs from its recent local baseline. \\
AS4 & level shift & 0.12 & segment & A persistent upward or downward change in a channel's baseline level is anomalous, but lower priority because the profile shows weak trend strength and limited autocorrelation rather than strong stable regimes. & Step-like increase near the change point followed by sustained elevation while the shifted level persists. \\
AS5 & correlation change & 0.08 & multivariate & A temporary change in cross-channel dependence is anomalous only when multiple channels move together or decouple in a way that contrasts with the normally low mean absolute correlation across 25 channels. & Elevated multivariate score over windows where cross-channel correlation structure changes. \\
\midrule
\multicolumn{6}{@{}l}{\emph{MITDB \#8 (univariate): one ECG channel, strong seasonality (period $\approx$ 25), strong autocorrelation}} \\
AS1 & contextual & 0.32 & contextual & Values or short runs that are unusual relative to their phase in the strong seasonal pattern, even if the raw magnitude is otherwise plausible; grounded in high seasonality\_strength 0.845, lag-1 autocorrelation 0.959, and dominant periodicity near 25 samples. & Peaks when observations depart from the expected seasonal phase/amplitude pattern and remains low on phase-consistent cycles. \\
AS2 & collective & 0.25 & segment & A sustained disruption of the normal cyclic structure, such as missing cycles, flattened oscillations, irregular cycle lengths, or a temporary loss of periodicity; grounded in strong spectral concentration, PD\_PeriodicityWang\_th0\_01 around 25 globally, and local periodicity variability. & Sustained high over the disrupted run, with scores returning to baseline when regular periodic structure resumes. \\
AS3 & point spike & 0.20 & point & Isolated extreme positive or negative excursions inconsistent with the heavy-tailed but autocorrelated baseline; grounded in high skewness 3.64, kurtosis 22.10, and outlier-related features indicating rare tail behavior. & Sharp peaks at isolated extreme samples with rapid decay before and after. \\
AS4 & level shift & 0.13 & segment & A persistent upward or downward change in the baseline or seasonal centerline beyond the weak global trend; grounded in low trend\_strength 0.032 but noticeable local histogram-mode changes across windows. & Step-like increase near the change point followed by sustained elevation while the shifted level persists. \\
AS5 & collective & 0.10 & segment & A sustained change in local variability or noise texture, such as a burst of roughness or unusually smooth behavior, while the mean seasonal pattern may remain present; grounded in high noise\_level 0.879 and local forecast-error/complexity feature variation. & Sustained high over the variance-regime interval, proportional to the persistence of abnormal roughness or smoothing. \\
\bottomrule
\end{tabular}%
}
\end{table*}

\subsection{End-to-End Run Trajectory}
\label{app:trajectory}

We trace one complete episode, from input to audited report, for MGAB \#7 (univariate), taken from the reported run with benchmark cards and without domain context. The time series has $100{,}000$ points in one channel and contains ten labeled anomalies of $20$ points each ($0.2\%$ of the time series). The labels are used only to score the outcome after the run. Table~\ref{tab:trajectory-example} summarizes every step and contrasts it with the run on the same time series without benchmark cards.

\noindent\textbf{Grounding.} The profiler reports a smooth, strongly periodic signal: seasonality $0.95$, lag-one autocorrelation $0.99$, near-zero trend, and a dominant period of $48$ samples. The grounding agent freezes a five-entry contract with $19$ validation questions. In priority order, the entries are AS1 contextual ($0.38$), \emph{``values or short runs that are unusual relative to their expected position in the strong 48-sample seasonal cycle''}; AS2 collective ($0.27$), a segment of one or more cycles whose waveform is \emph{``distorted, damped, phase-shifted, or loses its usual autocorrelation''}; AS3 point spike ($0.18$); AS4 level shift ($0.11$), which the contract itself ranks low \emph{``because trend\_strength and stationarity\_proxy are near zero''}; and AS5, a variance-regime change ($0.06$). Retrieval returns eight benchmark cards with confidence $0.93$. Five come from other MGAB training time series, and all five rank KNN and LOF as the top two detectors ($0.59$--$0.70$ VUS-PR), while no other detector exceeds $0.38$ on them. The remaining three are UCR sensor time series, one of which ranks IForest first.

\noindent\textbf{Validation protocol.} Before any candidate runs, the syllabus designer fixes three synthetic interventions (scale change, contextual, and collective, each at a $3\%$ rate) and a perturbed copy of the input. Validation runs on a common frame of $10{,}000$ points obtained by $10\times$ downsampling.

\noindent\textbf{Planning and refinement.} The budget allows four proposals, and the planner fills the contract in priority order:
\begin{itemize}[noitemsep,topsep=2pt,parsep=0pt,partopsep=0pt,leftmargin=0.4cm]
    \item \emph{Step 0 (KNN for AS1).} All entries are uncovered, so the planner targets AS1. It sets the window to the profiled period so that \emph{``distances compare full-cycle contextual morphology rather than isolated raw amplitudes''} and cites the cards for KNN. Validation gives AS1 a coverage of only $0.344$.
    \item \emph{Step 1 (LOF for AS2).} The planner moves to the next uncovered entry and explains the switch: \emph{``C1 used KNN for AS1 contextual point/short-run deviations and achieved only partial AS1 coverage. This proposal targets the still-uncovered AS2 [\dots] with a density-ratio detector rather than nearest-neighbor distance magnitude.''} LOF reaches the highest coverage of the episode ($0.851$).
    \item \emph{Step 2 (IForest for AS3).} For point spikes, the planner shortens the window because \emph{``a short window of 5 emphasizes local impulse shape and limits temporal smearing''} and cites the sensor card on which IForest ranks first.
    \item \emph{Step 3 (KMeansAD for AS4).} The planner acknowledges that \emph{``a persistent baseline offset is not strongly expected but remains an uncovered rubric hypothesis''}. Validation flags the candidate as unstable under perturbation ($0.50$).
\end{itemize}
The budget is exhausted before AS5 is reached. All four candidates pass the numeric sanity checks and enter the ensemble.

\noindent\textbf{Aggregation and outcome.} Each candidate targets a different entry, so each semantic group holds one candidate. The weights in Table~\ref{tab:trajectory-example} follow the validation proxies, and LOF receives the largest share. The quantile threshold flags $3{,}000$ points. All $200$ labeled anomaly points are among them, and every one is attributed to the AS2 group (LOF, $130$) or the AS1 group (KNN, $70$). The IForest and KMeansAD groups account for the remaining $2{,}136$ flagged points, none of them labeled. After the run, LOF and KNN alone reach $0.704$ and $0.674$ VUS-PR, IForest and KMeansAD below $0.01$, and the ensemble $0.684$. Because the final score takes the maximum across groups, the two uninformative groups add false alarms but do not displace the two informative ones, and they cost $0.02$ VUS-PR relative to LOF alone. The auditor checks the $12$ claims of the explanation: $9$ are supported, $1$ is partially supported, and $2$ are unsupported and removed. The episode takes $283$\,s and $88.2$K tokens.

\noindent\textbf{The same time series without benchmark cards.} The contract is nearly identical (the same five entry types, with AS2 at $0.30$). Guided by the profile alone, the planner proposes KMeansAD, DeepAnt, HBOS, and IForest for AS1--AS4, each with a rationale that is plausible for a periodic signal. Their coverages ($0.345$, $0.776$, $0.342$, $0.453$) are comparable to those of the run with cards, but no candidate exceeds $0.025$ VUS-PR. The ensemble reaches $0.021$, and only $18$ of the $200$ labeled points are flagged. The label-free coverage does not distinguish the two slates, because it measures how a detector responds to synthetic interventions, not to this time series' anomalies. What separates the runs is the first proposal, which the benchmark cards steer toward the detectors that worked on related time series. Validation then ranks candidates within the slate that planning produced. This matches the aggregate effect in Figure~\ref{fig:ensemble-size}, where the cards mainly improve the first candidate.

\noindent\textbf{The same time series with coding agents.} Neither coding agent turns its view of the anomalies into a check on its candidates. \emph{Claude Code} reads the catalog, runs a default IForest, and then switches to IForest on $100$-point windows because MGAB anomalies are ``subtle contextual insertions''. It validates only the output format and stops after one detector. That detector is the one \system{} assigned to point spikes, and both reach about $0.006$ VUS-PR ($194$\,s, $324$K tokens). \emph{Codex} explores more widely. It probes several catalog detectors, discards KNN as unstable after a native-library crash and LOF as too slow, and builds a rank-sum of $19$ signals: smoothing, median and derivative residuals, autoregressive residuals from ridge, Huber and extra-trees models fit on the first $20\%$ of the time series, an isolation forest on derived features, and TSB-AD IForest, HBOS, PCA, and KMeansAD with windows of $20$--$100$. The two discarded detectors are the ones that work here ($0.674$ and $0.704$ VUS-PR alone), and no included signal is checked against a stated anomaly type, so the sum reaches $0.010$ ($560$\,s, $673$K tokens). Both agents thus spend more tokens than \system{} without a criterion for which candidate should detect what.

\begin{table*}[t]
\centering
\caption{End-to-end trajectory on MGAB \#7 (univariate) from the reported run with benchmark cards and from the run without them (no domain context in either). Each step proposes one candidate for one contract entry. $\pi$ is the planner prior, $q$ the semantic coverage after validation, and $w$ the final ensemble weight. Flagged gives the thresholded points attributed to the candidate's semantic group, with the number inside the $200$ labeled anomaly points in parentheses. VUS-PR is computed post hoc from the candidate's own score; labels are never seen during the run.}
\label{tab:trajectory-example}
\footnotesize
\setlength{\tabcolsep}{4pt}
\renewcommand{\arraystretch}{1.1}
\begin{tabular}{@{}c l c l c c c c r c@{}}
\toprule
\textbf{Step} & \textbf{Target entry} & \textbf{Prio.} & \textbf{Detector} & \textbf{Window} & $\boldsymbol{\pi}$ & $\boldsymbol{q}$ & $\boldsymbol{w}$ & \textbf{Flagged (labeled)} & \textbf{VUS-PR} \\
\midrule
\multicolumn{10}{@{}l}{\emph{With benchmark cards} (final VUS-PR $0.6842$; $283$\,s; $88.2$K tokens)} \\
0 & AS1 contextual & 0.38 & KNN & 48 & 0.74 & 0.344 & 20.5\% & 310 (70) & 0.6742 \\
1 & AS2 collective & 0.27 & LOF & 48 & 0.72 & 0.851 & 30.0\% & 554 (130) & 0.7039 \\
2 & AS3 point spike & 0.18 & IForest & 5 & 0.64 & 0.497 & 23.9\% & 893 (0) & 0.0056 \\
3 & AS4 level shift & 0.11 & KMeansAD & 48 & 0.58 & 0.588 & 25.6\% & 1{,}243 (0) & 0.0051 \\
\midrule
\multicolumn{10}{@{}l}{\emph{Without benchmark cards} (final VUS-PR $0.0207$; $194$\,s; $80.5$K tokens)} \\
0 & AS1 contextual & 0.38 & KMeansAD & 48 & 0.72 & 0.345 & 26.5\% & 467 (0) & 0.0052 \\
1 & AS2 collective & 0.30 & DeepAnt & 100 & 0.74 & 0.776 & 17.4\% & 2{,}422 (16) & 0.0247 \\
2 & AS3 point spike & 0.16 & HBOS & 48 & 0.58 & 0.342 & 26.4\% & 0 (0) & 0.0046 \\
3 & AS4 level shift & 0.11 & IForest & 48 & 0.62 & 0.453 & 29.8\% & 111 (2) & 0.0056 \\
\bottomrule
\end{tabular}
\end{table*}

\subsection{Detection Report Example}
\label{app:report-example}

We reproduce the detection report for the SMAP time series of Table~\ref{tab:contract-examples} (SMAP \#17, multivariate) from the reported run with benchmark cards and without domain context. The run produced $60$ evidence records, four candidates, and a final ensemble of KNN, PCA, KMeansAD, and IForest.

\noindent\textbf{Report structure.} The report is rendered from the evidence bank in a fixed order:
\begin{itemize}[noitemsep,topsep=2pt,parsep=0pt,partopsep=0pt,leftmargin=0.4cm]
    \item \emph{Grounding}: the input profile (E1), the suspected anomaly types (E2), the eight retrieved benchmark cards (E3--E10), and the frozen contract with its five entries and $15$ validation questions (E12).
    \item \emph{Candidate generation}: for every planner step, the proposal record with the targeted contract entry, prior, rationale, expected signal, and preprocessing choice, followed by the pipeline that was built and its execution status and runtime.
    \item \emph{Validation}: the shared intervention syllabus (E17) and, for each candidate, its synthetic-injection metrics, perturbation stability, and the resulting validation feedback and proxy.
    \item \emph{Aggregation}: the final selection, threshold, and number of flagged points (E55), the contribution of each candidate (E56--E59), and the number of flagged timestamps attributed to each semantic group.
    \item \emph{Explanation and audit}: the templated explanation after auditing, the verdict for every claim, and an audit summary (E60).
\end{itemize}

\noindent\textbf{Evidence excerpts.} Representative records, quoted from the evidence bank:
\begin{itemize}[noitemsep,topsep=2pt,parsep=0pt,partopsep=0pt,leftmargin=0.4cm]
    \item \textbf{E1} (grounding, profiler). \emph{``Series has 7331 points, 25 channel(s). Mean abs cross-channel correlation=0.068, seasonality=0.147, trend=0.037, noise=0.729, stationarity\_proxy=0.422, lag1\_autocorr=0.059, skew=4.113, kurt=31.124.''}
    \item \textbf{E4} (grounding, benchmark memory). \emph{``BenchmarkCard 011\_MSL\_id\_10\_Sensor\_tr\_1525\_1st\_4590.csv [Sensor]: best detectors $\rightarrow$ PCA=0.90, KMeansAD=0.88, IForest=0.71.''}
    \item \textbf{E17} (validation, syllabus designer). \emph{``Validation syllabus [rubric] with 3 spec(s): IS1:spike@0.030(w=0.34), IS2:mean\_shift@0.030(w=0.34), IS3:contextual@0.030(w=0.34).''}
    \item \textbf{E19} (planning, first proposal; abridged). \emph{``Step 0: proposed IForest (prior=0.72). Target AS1 because all semantics have coverage 0.0 and AS1 has the highest priority tie-breaker (0.34). The profile shows strong right skew and very high kurtosis with weak lag-1 autocorrelation and low trend/seasonality, supporting isolated tail-spike morphology rather than sustained predictable structure (E1, E2). [\dots] Benchmark memory is mixed but includes IForest as a strong sensor baseline on a similar MSL card (E4) [\dots]''}
    \item \textbf{E46} (planning, fourth proposal; abridged). \emph{``Step 3: proposed KNN (prior=0.60). Target AS4 because it is tied for the lowest remaining semantic coverage at 0.0 [\dots] | why not previous: C1 IForest covered AS1 point spikes but had weak proxy detection; C2 KMeansAD covered AS2 collective windows but was perturbation-unstable; C3 PCA covered AS3 contextual residuals but had weak AP and instability. This proposal covers the still-uncovered AS4 level\_shift semantic with a neighbor-density mechanism rather than isolation, window clustering, or linear reconstruction.''}
    \item \textbf{E52} (validation feedback). \emph{``Validation feedback for C4 (KNN): final\_score=0.441, failure\_flags=[], reasons=[].''}
    \item \textbf{E55} (aggregation). \emph{``Selected candidates: [`C4', `C3', `C2', `C1']. Threshold=4.2128 via quantile, flagged=220 points.''}
    \item \textbf{E56} (aggregation). \emph{``Candidate C4 (KNN, targets=[`AS4']) contributes 27.8\% of the final ensemble (proxy=0.77, prior=0.18).''}
    \item \textbf{E60} (reporting, auditor). \emph{``Audited 12 claims: 8 supported, 2 partial, 2 unsupported, 0 contradicted.''}
\end{itemize}

\noindent\textbf{Tracing alerts.} Each candidate targets one contract entry, so each semantic group in Eq.~\eqref{eq:within-group} holds a single candidate: IForest for AS1 (point spike), KMeansAD for AS2 (collective), PCA for AS3 (contextual), and KNN for AS4 (level shift). Of the $220$ flagged timestamps, $154$ are driven by the AS3 group and $66$ by the AS1 group, so every alert traces to PCA or IForest. KNN carries the largest ensemble weight ($27.8\%$) yet drives no alert. With one candidate per group, the weight cancels in Eq.~\eqref{eq:within-group}, so each group score is that candidate's standardized score, and the maximum in Eq.~\eqref{eq:semantic-ensemble} does not depend on the weights. The weights would matter only in groups that pool several candidates. The report therefore records the weights and the alert attribution as separate quantities.

\begin{table*}[t]
\centering
\caption{Claim audit of the detection report for SMAP \#17 (multivariate), reproduced verbatim from the reported run except that inline citation markers are moved to the Cited column. Cited lists the evidence identifiers in the claim; Kept indicates whether the claim appears in the emitted explanation (\ding{51}: kept; $\sim$: kept with a \emph{partially supported} marker; \ding{55}: removed).}
\label{tab:report-audit}
\footnotesize
\setlength{\tabcolsep}{3pt}
\renewcommand{\arraystretch}{1.1}
\resizebox{\textwidth}{!}{%
\begin{tabular}{@{}l p{0.62\textwidth} l l c p{0.2\textwidth}@{}}
\toprule
\textbf{ID} & \textbf{Claim} & \textbf{Cited} & \textbf{Verdict} & \textbf{Kept} & \textbf{Auditor note} \\
\midrule
C1 & The input series has 7331 points across 25 channel(s) with seasonality strength 0.15, mean cross-channel correlation 0.07, and stationarity proxy 0.42. & E1 & supported & \ding{51} & -- \\
C2 & Anomaly semantics suggest these types: point\_spike, collective, contextual, level\_shift, correlation\_change. & E2 & supported & \ding{51} & -- \\
C3 & IForest was selected as a candidate anomaly semantic/score stream with final score 0.552 (synthetic AP=0.08, recall@k=0.02, perturbation stability=0.72). & E19, E28 & supported & \ding{51} & -- \\
C4 & KMeansAD was selected as a candidate anomaly semantic/score stream with final score 0.695 (synthetic AP=0.13, recall@k=0.12, perturbation stability=0.32). & E30, E36 & supported & \ding{51} & -- \\
C5 & PCA was selected as a candidate anomaly semantic/score stream with final score 0.728 (synthetic AP=0.09, recall@k=0.03, perturbation stability=0.50). & E38, E44 & supported & \ding{51} & -- \\
C6 & KNN was selected as a candidate anomaly semantic/score stream with final score 0.774 (synthetic AP=0.37, recall@k=0.34, perturbation stability=0.82). & E46, E52 & supported & \ding{51} & -- \\
C7 & The final ensemble threshold was 4.2128 and flagged 220 anomalous time points. & E55 & supported & \ding{51} & -- \\
C8 & The planner first proposed candidate models from the model-card library; semantic scorers were introduced only as a gated fallback for high-priority, under-covered semantics. & E19 & unsupported & \ding{55} & Fails the numeric check against E19. \\
C9 & Candidate contributions to the final ensemble: C4 (28\%), C3 (26\%), C2 (25\%). & E55 & unsupported & \ding{55} & Fails the numeric check against E55. \\
C10 & Dominant semantics at flagged points: AS3 (154), AS1 (66). & E55 & supported & \ding{51} & -- \\
C11 & Benchmark evidence supported the model choices: retrieved 8 card(s) (164\_SMAP\_id\_21\_Sensor\_tr\_1976\_1st\_4200.csv, 011\_MSL\_id\_10\_Sensor\_tr\_1525\_1st\_4590.csv, 002\_MSL\_id\_1\_Sensor\_tr\_500\_1st\_900.csv). & E3 & partially supported & $\sim$ & Citation valid; retrieved evidence matches weakly. \\
C12 & Limitations: synthetic-anomaly proxy validation may not reflect the true anomaly distribution, and the selection is unsupervised by design. & -- & partially supported & $\sim$ & Matched by retrieved evidence; no citation. \\
\bottomrule
\end{tabular}%
}
\end{table*}

\noindent\textbf{Audit.} Table~\ref{tab:report-audit} lists the twelve claims of the templated explanation and their verdicts. A claim is supported when its citations exist, its content matches the retrieved evidence, and at least one of its numbers agrees with a cited record within a $6\%$ relative tolerance. A claim with valid citations but weak topical match, or with a topical match but no citation, is partially supported. A claim whose numbers disagree with every cited record is unsupported, and a selection claim that names a rejected candidate is contradicted. In this report, eight claims are supported and kept. C11 and C12 are kept with a marker. C8 and C9 are removed. C9 is correct in substance, as E56--E58 give contributions of $27.8\%$, $26.4\%$, and $25.3\%$, but it cites E55, which stores the contributions as fractions rather than percentages; the auditor therefore cannot match it and drops it. The audit is conservative in this sense: it removes claims it cannot match to their cited records, including some correct ones, and it keeps only claims that are consistent with the records they cite. In claims C3--C6, the number labeled ``final score'' is the candidate's validation proxy, which also appears in E56--E59; the separately recorded validation-feedback score (e.g., $0.441$ for KNN in E52) is a different quantity.

\noindent\textbf{Emitted explanation.} After the audit, the report emits C1--C7 and C10--C12 in order, with C11 and C12 followed by ``[partially supported]''. The explanation states what the run did and on which evidence; it does not assert that the flagged points are true anomalies. On this time series the run reaches a VUS-PR of $0.3135$ (Appendix~\ref{app:domain-context}).

\subsection{Domain Context Examples}
\label{app:domain-context}

\noindent\textbf{Aggregate effect.} Context leaves $36$ of $115$ univariate and $17$ of $31$ multivariate time series unchanged. It also helps the two-lead multivariate ECGs. It adds little for the heterogeneous UCR archive, whose description can say little beyond ``no trusted units''. The univariate Human Activity and Sensor time series where it hurts are single channels cut out of the multichannel OPPORTUNITY and SMAP recordings; the gate bounds the harm of their uninformative descriptions.

\noindent\textbf{Construction.} Each source dataset receives one short description per benchmark setting, curated from the public TSB-AD source catalogue. A description may state what the signal measures, the roles of its channels, which variations are normal, and which evidence the representation cannot support. It is written from the source identity and the channel schema only: labels, the training-split boundary (\texttt{tr\_}), and the first-anomaly index (\texttt{1st\_}) in TSB-AD file names are never used. The same source can receive different descriptions in the two settings, because the univariate time series expose one extracted channel. Representative descriptions follow.

\begin{itemize}[noitemsep,topsep=2pt,parsep=0pt,partopsep=0pt,leftmargin=0.4cm]
    \item \textbf{SMAP, multivariate.} \emph{``Spacecraft telemetry schema: channel 0 is the continuous measurement and channels 1..24 are binary command/state inputs. Remove inactive command bits, never score commands as targets, and test target response to current/lagged command state with a one-sided transition grace interval. Keep target-only persistence as a separate hypothesis.''}
    \item \textbf{SMAP, univariate.} \emph{``Univariate spacecraft telemetry target extracted without the command/state channels available in the multivariate representation. Test target-only persistent deviations and local regime changes. Do not claim command-response inconsistency because the conditioning channels are absent.''}
    \item \textbf{MITDB, multivariate.} \emph{``Two-lead ambulatory ECG. Ordinary sharp QRS peaks are recurring normal morphology. Estimate beat anchors from the observed leads, align beats before building a robust template, score beat-morphology and RR-interval deviations, and use cross-lead coherence as separate evidence; do not use a fixed row-count seasonal phase template.''}
    \item \textbf{SVDB, univariate.} \emph{``Single-lead supraventricular-arrhythmia ECG signal. Ordinary sharp QRS complexes are recurring normal morphology rather than point anomalies by amplitude alone. Use feature-derived beat alignment, morphology residual, and RR-interval deviation; cross-lead coherence is unavailable.''}
    \item \textbf{OPPORTUNITY, univariate.} \emph{``Univariate projection of wearable or object activity sensing. Legitimate activity transitions may change level and variance abruptly. With no activity annotation or companion channels exposed, use conservative local persistence and robust-deviation evidence on the sole target.''}
    \item \textbf{UCR, univariate.} \emph{``Univariate UCR anomaly-archive signal drawn from heterogeneous application domains. Source identity alone does not establish units, bounds, or a physical operating model. Use the single signal as the target and prefer feature-derived multiscale residual, robust-tail, and regime-change evidence.''}
    \item \textbf{TODS, univariate.} \emph{``Synthetic univariate anomaly-detection signal with no trusted physical semantics. Use only observed temporal structure: robust local residual, isolated tail deviation, and sustained level or variance change. Do not infer units or hard physical bounds.''}
\end{itemize}

\noindent\textbf{From description to scorer.} Given the description, the time series profile, and the frozen no-context result, the LLM states at most two domain hypotheses. Each hypothesis names an invariant, what it adds to the frozen ensemble, and the alternatives it rejected, and it is compiled into a whitelisted semantic scorer with no LLM-chosen numerical parameters. The scorer passes through the same synthetic-intervention validation as any other candidate and joins the ensemble only if it clears the label-free benefit gate. Two cases illustrate the extremes of Figure~\ref{fig:domain-context}.

\emph{A gain: SMAP time series \#17, multivariate} (VUS-PR $0.3135\rightarrow0.3541$). From the multivariate SMAP description, the LLM proposed the invariant \emph{``telemetry channel 0 should follow a context-conditioned response distribution for the active binary command-state channels''}. It noted that the frozen ensemble (KNN, PCA, KMeansAD, IForest) covers general point, collective, and contextual structure but never conditions the target on command state, and it rejected a lagged-dependency scorer because the schema supports state-conditioned rather than lagged responses. The compiled contextual scorer localized well in validation (localization $0.997$, normal-region false-positive score $0.992$) and was admitted to the ensemble.

\emph{A loss: OPPORTUNITY time series \#25, univariate} (VUS-PR $0.9984\rightarrow0.8092$). The univariate description offers no positive structure to test; it mostly warns that activity transitions legitimately change level and variance. Guided by the time series profile (lag-one autocorrelation $0.978$), the LLM proposed that \emph{``short bursts of elevated roughness or local variance on the sole target channel are anomalous''} and compiled a rolling-variance scorer. The scorer passed the gate, but its high scores coincide with exactly the activity transitions the description calls normal, which diluted an already near-perfect no-context ensemble. The case shows that the gate bounds, but does not eliminate, the cost of a description that cannot be turned into a specific test.

\end{document}